\documentclass[journal]{IEEEtran}
\usepackage{amsmath,graphicx,amssymb, amsthm, algorithm, bm,epstopdf,subcaption,cite,multirow}
\usepackage[noend]{algpseudocode}
\usepackage{enumitem}
\usepackage{pstricks,pst-plot, pst-node}
\PassOptionsToPackage{pdf}{pstricks} 
\usepackage{hyperref}
\usepackage{footnote}
\usepackage{comment}

\newenvironment{talign*}
 {\let\displaystyle\textstyle\csname align*\endcsname}
 {\endalign}

\newenvironment{tflalign*}
 {\let\displaystyle\textstyle\csname flalign*\endcsname}
 {\endalign}

\newtheorem{definition}{Definition}
\newtheorem{theorem}{Theorem}

\newtheorem{lemma}{Lemma}
\newtheorem{proposition}{Proposition}
\newtheorem{remark}{Remark}

\DeclareMathOperator*{\argmin}{argmin} 
\DeclareMathOperator*{\argmax}{argmax} 
\DeclareMathOperator{\sign}{sign}
\newcommand{\mt}[1]{\mathbf{#1}} 
\newcommand{\vt}[1]{\bm{\mathrm{#1}}} 
\newcommand{\test}[2]{\mu(#1|#2)} 
 
\newcommand{\proj}[2]{\mathrm{proj}_{#2} #1} 

\usepackage{comment} 

\begin{document}

\title{Stable safe screening and structured dictionaries for faster $\ell_{1}$ regularization}

\author{C\'{a}ssio~F.~Dantas,~\IEEEmembership{Student Member,~IEEE,}
        and R\'{e}mi~Gribonval,~\IEEEmembership{Fellow,~IEEE}% <-this % stops a space
\thanks{The authors are with Univ Rennes, Inria, CNRS,  IRISA, F-35000 Rennes, France (e-mail: cassio.fraga-dantas@inria.fr). This work has been partially funded by the Becose ANR project (ANR-15-CE23-0021).}%
\thanks{Code is available at: \href{https://github.com/cassiofragadantas/Screening_ADST}{github.com/cassiofragadantas/Screening\_ADST}~\cite{F.Dantas2019a}.}
}

\maketitle

\begin{abstract}

In this paper, we propose a way to 
combine two acceleration techniques for the $\ell_1$-regularized least squares problem: safe screening tests, which allow to eliminate useless dictionary atoms; and the use of fast structured approximations of the dictionary matrix.
To do so, we introduce a new family of screening tests, termed \emph{stable screening}, which can cope with approximation errors on the dictionary atoms while keeping the safety of the test 
(i.e. zero risk of rejecting atoms belonging to the solution support). 
Some of the main existing screening tests are extended to this new framework.
The proposed algorithm consists in using a coarser (but faster) approximation of the dictionary at the initial iterations and then switching to better approximations until eventually adopting the original dictionary.
A systematic switching criterion based on the duality gap saturation and the screening ratio is derived.
Simulation results show
significant 
reductions in both computational complexity and execution times for a wide range of tested scenarios.

\end{abstract}

\IEEEpeerreviewmaketitle

\section{Introduction}

Sparsity-constrained linear regression has found numerous applications in signal processing and machine learning, tackling under-determined inverse problems. 
These appear in many forms such as image or audio inpainting \cite{Adler2012}, source localization \cite{Malioutov2005} or spike deconvolution \cite{Dossal2005}, to cite only a few.

There are many computational approaches to estimate the sparse coefficient vector in such settings with two main families: 
greedy algorithms \cite{Mallat1993}~\cite{Pati1993},
and convex optimization approaches relying on $\ell_{1}$-norm minimization, where the regression problem is expressed as an $\ell_1$ regularized optimization problem --known as Basis Pursuit \cite{Chen1998} or LASSO \cite{Tibshirani1996}-- which solution is computed using iterative convex optimization algorithms such as ISTA or its variants \cite{Daubechies2004,Beck2009,Bioucas-Dias2007,Wright2009,Chambolle2011}.

For very high-dimensional problems, however, iterative algorithms to solve $\ell_{1}$ minimization problems can become computationally prohibitive, which is why accelerating techniques are still an intense research topic.
This paper demonstrates how to combine two such techniques:
\begin{enumerate}
\item Fast structured operators \cite{Rubinstein2010a,Chabiron2015, Magoarou2016,Sulam2016,Dantas2017} which provide faster matrix-vector products 
(see Section \ref{ssec:approx_safe_region});
\item Safe screening tests \cite{ElGhaoui2010,Xiang2011,Bonnefoy2015,Fercoq2015,Ndiaye2017,Xiang2016}, which safely eliminate unused explanatory variables
(see Section \ref{sec:screening}).
\end{enumerate}

This paper extends the results in \cite{F.Dantas2017} \cite{F.Dantas2018}. 
In \cite{F.Dantas2017} we introduced a safe screening sphere test that manipulates an approximate dictionary 
and in \cite{F.Dantas2018} the GAP Safe sphere test \cite{Fercoq2015} was extended to this new setting.
Here, we revisit the previous results under a broader formalism and propose a fast algorithm for $\ell_1$-minimization which combines safe screening and (potentially multiple) fast approximate dictionaries.

\subsection{Sparsity constrained regularization}

Let $N$ and $K$ be  respectively the dimension of the observed vector and that of the unknown coefficient vector. 
The observed vector is denoted $\vt{y} \in \mathbb{R}^N$
and modeled as $\vt{y} \approx \mt{A} \vt{x}_{0}$ where $\vt{x}_{0} \in \mathbb{R}^{K}$ is sparse and ${\mt{A} = [\vt{a}_1, \dots, \vt{a}_K]} \in \mathbb{R}^{N\times K}$. In the context of linear inverse problems, $\mt{A}$ is the so-called measurement matrix \cite{Foucart2013}, while for sparse signal representations $\mt{A}$ would be the dictionary matrix \cite{Chen1998}, and in statistics the design matrix \cite{Tibshirani1996}. In this paper, we adopt the terminology of sparse signal representations hence $\mt{A}$ is called a dictionary and its columns, denoted $\vt{a}_j\in \mathbb{R}^N$, are called atoms. 

The $\ell_1$-regularized least squares, referred to as Lasso or Basis Pursuit,
 consists in finding a sparse coefficient vector $\vt{x} \in \mathbb{R}^K$, solution of the following optimization problem:
\begin{align} \label{eq:lasso}
\mathcal{L}(\lambda,\mt{A},\vt{y}): & &
\vt{x}^\star  = {\argmin_{\vt{x}}}  \underbrace{\frac{1}{2} \|\mt{A}\vt{x} - \vt{y}\|_2^2 + \lambda \| \vt{x} \|_1}_{P(\vt{x}|\mt{A})}
\end{align}
where $P(\vt{x}|\mt{A})$ is called the primal objective and the parameter $\lambda \!>\! 0$ controls the trade-off between data fidelity and sparsity of the solution.
We suppose that 
\begin{equation}
\lambda \leq \lambda_{\max}:=\|\mt{A}^T\vt{y}\|_\infty \label{eq:DefLMax}
\end{equation}
since otherwise $\vt{x}^\star = \vt{0}\in \mathbb{R}^K$ is the unique solution.

\subsection{Iterative algorithms}
First-order iterative algorithms, especially proximal-gradient methods (ISTA~\cite{Daubechies2004}, FISTA~\cite{Beck2009}, TwIST~\cite{Bioucas-Dias2007}, SpaRSA~\cite{Wright2009}, Chambolle-Pock~\cite{Chambolle2011}), 
are popular approaches for solving \eqref{eq:lasso}.

One can use an abstract notations to represent the update step for a generic iterative algorithm for the problem $\mathcal{L}(\lambda,\mt{A},\vt{y})$:
$\{\vt{x}_{t+1},\vt{\alpha}_{t+1}\} \gets p(\mt{x}_{t},\mt{A},\vt{\alpha}_{t})$
where $\mt{x}_{t}$ is the current estimate of the primal variable and $\vt{\alpha}_t$ is a list of updated auxiliary scalars (e.g. the gradient step-size, and possibly a few previous estimates of the primal variable). 
For instance, the ISTA algorithm is given by
\begin{align} \label{eq:ISTA_update}
\vt{x}_{t+1} \gets  p(\vt{x}_{t},\mt{A},L_t) 
\!=\! \operatorname{ST}_{\frac{\lambda}{L_t}}\left(\vt{x}_{t} +  \frac{1}{L_t}\mt{A}^T (\vt{y} - \mt{A}\vt{x}_{t}) \right)
\end{align}
where $\operatorname{ST}_u(z)\!=\! {\sign(z)(|z|-u)_{{}+{}}}$ denotes the soft-thresholding operation, which is the proximal operator associated to the $\ell_1$-norm.

\subsection{Computational bottlenecks}
The main bottleneck of existing iterative algorithms in terms of computational complexity is the cost of the required matrix-vector products involving the dictionary matrix, which dominates the overall iteration cost.
For example, ISTA requires two matrix-vector multiplications at each iteration (or at least one if the Gram matrix $\mt{A}^T\mt{A}$ is pre-computed and stored).

\subsection{Addressing the computational bottlenecks}
A popular way to address this limitation is to constrain the dictionary matrix to a certain type of structure which would allow for fast matrix-vector products.  Another approach is to use screening rules to identify 
a set indexing zero components of the solution $\vt{x}^\star$ {\em before even computing it}.

\subsubsection{Acceleration with structured dictionaries} \label{ssec:structured_dict}

Different types of structure, more or less suited to a given application, can be imagined. 
In this kind of work, there is always a compromise between the flexibility (generality) of the structure and the provided speedup.

In \cite{Rubinstein2010a}, the authors constrain the dictionary to be the product of a fixed \textit{base dictionary} that has a fast implementation (a DCT, for instance) and a column sparse matrix.
This model is extended in \cite{Sulam2016} by replacing the base dictionary with cropped wavelets, which have a certain degree of adaptability. 
A more general model, which has the two previous ones as particular cases, was proposed in \cite{Magoarou2016}.
The dictionary is constrained to be the product of several sparse matrices.

Another interesting line of research, particularly suited to multi-dimensional signals, consists in finding dictionaries formed as the Kronecker (or tensorial) product of sub-dictionaries of smaller size. It was proposed in \cite{Hawe2013} for the Kronecker product of two sub-dictionaries aiming 2D signals (like images). The model was then extended to a sum of Kronecker products in \cite{Dantas2017} and \cite{F.Dantas2018a} (respectively for 2-dimensional and higher-order tensorial data), increasing its flexibility in exchange of some additional complexity.

In practical applications, the dictionary matrix $\mt{A}$ is not necessarily structured. A possible strategy is to replace certain iterations $\vt{x}_{t+1} = p(\mt{x}_{t},\mt{A},\vt{\alpha}_{t})$ with $\vt{x}_{t+1} = p(\vt{x}_{t},\tilde{\mt{A}},\vt{\alpha}_{t})$ where $\tilde{\mt{A}}$ is a structured approximation of $\mt{A}$, i.e.,  
\begin{align} \label{eq:approx_dict}
\mt{A} = \tilde{\mt{A}} + \mt{E}
\end{align}
with approximation error  $\mt{E} \!=\! [\vt{e}_1, \dots, \vt{e}_K] \!\in\! \mathbb{R}^{N\times K}$.

\subsubsection{Acceleration with screening rules} \label{ssec:accel_with_screening}

For a given 
instance $\mathcal{L}(\lambda,\mt{A},\vt{y})$
of \eqref{eq:lasso}, characterized by a  regularization parameter $\lambda$, a dictionary $\mt{A}$, and an input vector $\vt{y}$, the safe screening approach, introduced by \cite{ElGhaoui2010}, consists in identifying and removing from the dictionary a subset of atoms which are guaranteed to have zero weight in a solution $\vt{x}^\star$, before solving the problem. By removing these so-called \emph{inactive atoms} a more compact and readily solved problem is obtained, with decreased matrix-vector multiplication cost, while not affecting at all the original solution which can be obtained by simply zero-padding its restricted version.

Put differently, safe screening is a feature selection technique for a given instance of \eqref{eq:lasso}. However, unlike other previous feature selection heuristics \cite{Fan2008,Tibshirani2011} based 
on correlation measures between the atoms $\vt{a}_j$ and the input signal $\vt{y}$, safe screening has zero risk of false rejections.
Moreover, screening techniques are transparent to the underlying $\ell_{1}$ solver and can be readily combined with almost any existing solvers.

Basically, three categories of screening rules can be distinguished: 1) Static; 2) Dynamic; 3) Sequential. The two first categories being mutually exclusive, but not the third one.

The earliest safe screening techniques \cite{ElGhaoui2010} \cite{Xiang2011} \cite{Xiang2012}, currently classified as \emph{static rules}, where designed to be applied once and for all before starting the optimization. 
In contrast, the more recent \emph{dynamic rules}, introduced in \cite{Bonnefoy2015} and followed by \cite{Fercoq2015}, are repeatedly applied during the iterative optimization algorithm leveraging its current solution estimate and gradually increasing the set of rejected atoms.

\emph{Sequential rules} \cite{Wang-Wonka2015,Malti2016,Xiang2011,Liu2014}
exploit the fact that \eqref{eq:lasso} 
is commonly solved over a grid of regularization parameters $\lambda_{i}$ and reuse the results from a previous configuration $\lambda_{i-1}$ to improve the screening performance.

\subsection{Contributions}

First, we introduce a formalism for  defining 
safe screening tests which are robust to approximation errors on the dictionary matrix given that an error bound is provided.
The resulting tests are called \emph{stable screening} tests.
The stable tests proposed here are general and apply to any context in which the atoms are known up to a certain error margin. The source of this error can be manifold, but we will focus on the case where it is a side effect of manipulating structured approximations of the true dictionary matrix. 
The proposed framework is also general with respect to the form of the safe region (be it a sphere, dome, etc) as well as to that of the error bound. Here, again, we stick to $\ell_p$-balls, which allows us to exemplify the stable tests on some existing $\ell_2$-sphere tests --one of which being the state-of-the-art GAP Safe \cite{Fercoq2015}-- for a general $\ell_q$-ball error bound.
Extension to dome tests is not particularly difficult within the proposed formalism.

In a second part, we exploit the proposed stable screening in a fast algorithm for $\ell_1$-minimization problems, by making use of fast structured approximations of the problem's dictionary matrix.
It consists in starting the iterative optimization by using a coarser (but faster) version of the true dictionary 
and, as approaching convergence, finer approximations are progressively adopted until eventually the original dictionary takes over.
Choosing an appropriate moment to switch to a more precise dictionary is crucial in the proposed algorithm. A robust switching criterion based on both the  duality gap saturation and the screening ratio is proposed.

\subsection{Related Work}

Apart from the aforementioned structured dictionaries and safe screening tests --as well as other preceding correlation-based feature selection heuristics \cite{Fan2008,Tibshirani2011}--
some related acceleration strategies for sparsity-inducing optimization problems (or more specifically for problem \eqref{eq:lasso}) are worth citing.

Instead of starting from the full problem and pruning the feature set, working set techniques \cite{Kim2010,Johnson2015} start with small restricted problems and progressively include more promising features. 
In \cite{Massias2017}, the authors combine a working set strategy with safe screening and in \cite{Massias2018} they incorporate a dual extrapolation technique to further enhance the screening performance and accelerate convergence. 
This idea is conceivably complementary to the techniques proposed here.

Joint screening \cite{Herzet2018} allows to screen many atoms which lie close together in one single test, reducing the number of required tests for a given dictionary. Interestingly, the resulting tests share many similarities and mathematical connections to the stable screening tests introduced here,
despite arising from an essentially different premise.

\subsection{Outline of the Paper}

This paper is organized as follows. In Section \ref{sec:screening} we briefly review standard (static and dynamic) safe screening rules. {\em Stable} safe screening rules are introduced in Section \ref{sec:screening_approx}. 
A complete description of the proposed algorithm is given in Section \ref{sec:algorithm} and experimental results in Section \ref{sec:experiments}.

\section{Reminders on Safe Screening} \label{sec:screening}

Given a set  $\mathcal{A}$ of non-repeating integers,
${\mt{A}_{[\mathcal{A}]}:= [\vt{a}_i]_{i \in \mathcal{A}}}$ denotes a sub-matrix of $\mt{A}$ composed of the columns indexed by the elements in $\mathcal{A}$. 
The notation extends to vectors: $\vt{x}_{[\mathcal{A}]} := [\vt{x}(i)]_{i \in \mathcal{A}}$, where $\vt{x}(i)$ denotes the $i$-th component of $\vt{x}$. Screening rules allow to identify such sets of reduced cardinality $|\mathcal{A}_{t}| < K$, so that the complexity of applying $\mt{A}_{[\mathcal{A}]}$ or its transpose is reduced compared to that of applying $\mt{A}$, while ensuring that $\mathcal{A}_{t}$ contains the support of the solution $\{i: \vt{x}^\star(i) \neq 0\}$.
This is achieved using the dual of \eqref{eq:lasso}~\cite{ElGhaoui2010}:

\begin{align} \label{eq:lasso_dual}
\vt{\theta}^{\star}  ={} & \argmax_{\vt{\theta} \in \Delta_{\mt{A}}} ~ \underbrace{\frac{1}{2}\|\vt{y}\|_2^2 - \frac{\lambda^2}{2}\left\Vert \vt{\theta} - \frac{\vt{y}}{\lambda}\right\Vert_2^2}_{D(\vt{\theta})} 
\end{align}
where 
\begin{align} \label{eq:dual_feasible_set}
\Delta_{\mt{A}} &\!=\! \{\vt{\theta} \!\in\! \mathbb{R}^{N}: ~ \|\mt{A}^T\vt{\theta}\|_\infty \!\leq\! 1\}
\end{align}
is the dual feasible set and $D(\vt{\theta})$ is the dual objective.
The dual and primal solutions ($\vt{\theta}^{\star}$ and  $\vt{x}^\star$) are linked through the relation $\vt{y} = \mt{A} \vt{x}^\star + \lambda \vt{\theta}^{\star}$.
Optimality conditions (KKT) at the dual solution $\vt{\theta}^{\star}$ read
(see for instance \cite{Xiang2016} for more details)
\[ 
\vt{a}^T_j \vt{\theta}^{\star} = \left\{ \begin{array}{lcc}
\sign(\vt{x}^\star(j)) & \text{if} & \vt{x}^\star(j) \neq 0 \\
\gamma \in [-1,1] & \text{if} & \vt{x}^\star(j) = 0
\end{array} \right.
, \quad \forall j \in  \{1,\dots,K\}.
\] 
Hence, every dictionary atom for which $|\vt{a}_j^T \vt{\theta}^{\star}| \!<\! 1$ is inactive. 

\subsection{Notion of safe region}
Since the optimal solution $\vt{\theta}^\star$ of the dual problem \eqref{eq:lasso_dual} is not known, the inner products $\vt{a}_j^T \vt{\theta}^\star$ cannot be evaluated. Fortunately, given only $\mt{A}$ and $\vt{y}$, it is possible to identify at a moderate computational cost a region 
$\mathcal{R}\subset \mathbb{R}^N$, 
called \emph{safe region}, which is guaranteed to contain the optimal $\vt{\theta}^\star$.

\begin{definition}[Safe region] \label{def:safe_region}
A region $\mathcal{R} \subset \mathbb{R}^N$ is safe (with respect to the dual problem \eqref{eq:lasso_dual}) if and only if it contains the dual solution $\vt{\theta}^\star$, i.e. $\vt{\theta}^\star\in \mathcal{R}$.
\end{definition}

\subsection{Screening an atom given a safe region}
Consider an atom $\vt{a}_{j}$. If the inequality $|\vt{a}_j^T \vt{\theta}| < 1$ holds for all $\vt{\theta} \in \mathcal{R}$ where $\mathcal{R}$ is a safe region, then the above analysis ensures that $\vt{a}_j$ is inactive. This gives rise to a {\em screening test}. 

\begin{definition}[Screening test for an atom] \label{def:screening_test}
Given a region $\mathcal{R}$, a screening test 
for the atom $\vt{a} \in \mathbb{R}^N$ is given by:
\begin{align}\label{eq:DefBasicTest}
\test{\vt{a}}{\mathcal{R}} = \sup_{\vt{\theta} \in \mathcal{R}}  |\vt{a}^T \vt{\theta}|
\end{align}
\end{definition}

A sufficient condition for an atom $\vt{a}_j$ to be inactive can be expressed as follows: if $\mathcal{R}$ is safe then
\begin{align*}
\test{\vt{a}_j}{\mathcal{R}} < 1 \quad \implies \quad \vt{x}^\star(j) = 0.
\end{align*}
In practice, for each atom $\vt{a}_j$, computing the test $\test{\vt{a}_j}{\mathcal{R}}$ allows to eliminate or not the atom. 

Formally, given a safe region, the atoms can be partitioned into a preserved set $\mathcal{A} = \{j \in \{1,\dots,K\} : \test{\vt{a}_j}{\mathcal{R}} \geq 1 \}$ and its complement, the rejection set $\mathcal{A^{\mathsf{c}}}$, that gathers the indices of the identified inactive atoms.

In practice, safe regions have simple parameterized shapes which parameters need to be identified with moderate computations from the only knowledge of $\mt{A}$, $\vt{y}$, and possibly the current iterate $\vt{x}_{t}$ of an iterative algorithm addressing \eqref{eq:lasso}.

The two most common shapes of safe regions in the literature are spheres and domes (i.e. intersection between a sphere and one or more half-planes).

This paper focuses on sphere tests. Assume we are given a safe region $\mathcal{R}$ which is a closed $\ell_p$-ball with center $\vt{c}$ and radius $R$, denoted $B_p(\vt{c},R) = \{\vt{z} : \|\vt{z} - \vt{c}\|_p \leq R\}$. The screening test for this region has a closed form (a proof is given in Appendix \ref{apx:sphere})
\begin{equation} \label{eq:sphere_test}
\test{\vt{a}}{B_p(\vt{c},R)} = |\vt{a}^T \vt{c}| + R\|\vt{a}\|_{p^*}
\end{equation}
where $\|\cdot\|_{p^*}$ denotes the dual norm associated to the $p$-norm,  with $\frac{1}{p}+ \frac{1}{p^*}=1$.
For simplicity, we omit subscripts for the $\ell_2$-ball ($p=p^*=2$) denoted $B(\vt{c},R)$.

\subsection{Construction of a safe spherical region} \label{ssec:safe_regions}

A safe region should be as small as possible (to maximize the screening effect) while requiring as little computational overhead as possible. In light of~\eqref{eq:sphere_test}, as $\|\vt{a}\|_{p^{*}}$ can be precomputed for each atom, the computational overhead of a screening test with a spherical region is  governed by the cost of computing the radius $R$ and the inner products $\vt{a}^{T}\vt{c}$ for all atoms that have not been screened out yet. This calls for techniques where the choice of $R$ and $\vt{c}$ allows to reuse either quantities that have already been computed along previous iterations of the optimization algorithm. 

The construction of the first (static) safe region, obtained in \cite{ElGhaoui2010}, follows from the simple observation that the solution $\vt{\theta}^\star$ of the dual problem \eqref{eq:lasso_dual} is the Euclidean projection of $\vt{y}/\lambda$ on the feasible set $\Delta_{\mt{A}}$. 
As a result, if some feasible point $\vt{\theta}_F \in \Delta_{\mt{A}}$ is known, then $\vt{\theta}^\star$ can't be further away from $\vt{y}/\lambda$ than $\vt{\theta}_F$ in the $\ell_2$ sense, 
i.e. $\|\vt{\theta}^\star - \vt{y}/\lambda\|_2 \leq \|\vt{\theta}_F - \vt{y}/\lambda \|_2 ~\forall \vt{\theta}_F \in \Delta_{\mt{A}}$.

This leads to an $\ell_2$-spherical safe region ($p\!=\!p^*\!=\!2$),
\begin{align} \label{eq:basic_sphere}
\mathcal{R} = B(\vt{c} = \vt{y}/\lambda, R = \|\vt{\theta}_F - \vt{y}/\lambda \|_2),
\end{align}
which full specification requires a feasible point $\vt{\theta}_F \in \Delta_{\mt{A}}$.

To generate a feasible point $\vt{\theta}_F \in \Delta_{\mt{A}}$, one could compute the Euclidean projection of any given point $\vt{z} \in \mathbb{R}^N$ over the closed convex set $\Delta_{\mt{A}}$. 
As this is too computationally demanding, simpler scaling approaches are preferred:
given $\vt{z}\in\mathbb{R^N}$ we determine a scaling factor $\alpha$ so that $\Theta(\vt{z}|\mt{A}) := \alpha \vt{z} \in \Delta_{\mt{A}}$. 

By definition of $\Delta_{\mt{A}}$, if $\|\mt{A}^T \vt{z} \|_\infty \leq 1$ no scaling is needed, otherwise one only needs to divide $\vt{z}$ by $\|\mt{A}^T \vt{z} \|_\infty$. This yields 

\begin{align} \label{eq:conditional_dual_scaling}
\Theta(\vt{z}|\mt{A}) := \frac{\vt{z}}{\max\left(1, \|\mt{A}^T \vt{z} \|_\infty \right)}.
\end{align}

While the scaling in \ref{eq:conditional_dual_scaling} selects a feasible point to build a safe sphere~\eqref{eq:basic_sphere} given an arbitrary $\vt{z}$, it neglects the fact that one would like to minimize the {\em radius} of this sphere to maximize the effect of screening. 
This leads to the \emph{dual scaling}~\cite{ElGhaoui2010} formalized as follows (a proof is given in \cite[Lemma 8]{Bonnefoy2015})
\begin{proposition}
Denoting $\proj{\dfrac{\vt{y}}{\lambda}}{\vt{z}} = \dfrac{\vt{y}^T \vt{z}}{\lambda \|\vt{z}\|_2^2} \vt{z}$ the Euclidean projection of  $\vt{y}/\lambda$ onto the direction of $\vt{z}$, we have
\begin{align*}
\Theta\left(\proj{\dfrac{\vt{y}}{\lambda}}{\vt{z}}  ~\vert ~ \mt{A} \right) = \argmin_{\vt{z}'= \alpha \vt{z}} \|\vt{z}' - \vt{y}/\lambda\|_2 \quad \text{s.t. } \|\mt{A}^T\vt{z}'\|_{\infty} \leq 1
\end{align*}
i.e. among all points  which are both dual feasible and proportional to $\vt{z}$, $\Theta\left(\proj{\dfrac{\vt{y}}{\lambda}}{\vt{z}} ~\vert ~ \mt{A} \right)$ is the one closest to $\vt{y}/\lambda$.
\end{proposition}

\

To obtain dynamic spherical regions, we need dual feasible points that change over iterations ($\vt{\theta}_{F} = \vt{\theta}_t$ at iteration $t$) and leverage the current state of the optimization algorithm. Following \cite{Bonnefoy2015}, this can be achieved by defining $\vt{\theta}_t$ proportional to the current residual $\vt{\rho}_t := \vt{y} - \mt{A}\vt{x}_t$,
since the dual solution $\vt{\theta}^\star$ is proportional to $\vt{y} - \mt{A}\vt{x}^\star$.
Taking $\vt{z} = \vt{\rho}_t$ 
and
$\vt{\theta}_t = \Theta\left(\proj{\dfrac{\vt{y}}{\lambda}}{\vt{\rho}_t}  ~\vert ~ \mt{A} \right)$
yields 
\begin{align} \label{eq:theta_dynamic}
\vt{\theta}_t 
=& \left[\dfrac{\vt{y}^T \vt{\rho}_t}{\lambda \|\vt{\rho}_t\|_2^2}\right]_{-\alpha_t}^{~ \alpha_t} \vt{\rho}_t, 
& \text{with }  &
\alpha_t = ({\|\mt{A}^T\vt{\rho}_t \|_\infty})^{-1}
\end{align}
where $[z]_a^b:= \min(\max(z,a),b)$ denotes the projection of the scalar $z$ onto the segment $[a,b]$.

The above principles yield existing safe spherical regions.
\subsubsection{Static Safe sphere \cite{ElGhaoui2010}}

With $\vt{z} = \vt{y}/\lambda$ in \eqref{eq:conditional_dual_scaling} one gets  $\theta_{F} = \Theta(\vt{y}/\lambda | \mt{A}) = \vt{y}/\lambda_{\max}$, and~\eqref{eq:basic_sphere} reads 
$\mathcal{R} = B(\vt{c},R)$ with
\begin{align}\label{eq:static_sphere}
\text{Static Safe: }  \quad 
\vt{c}=\vt{y}/\lambda, R=|\tfrac{1}{\lambda_{\max}} - \tfrac{1}{\lambda}| \cdot \|\vt{y}\|_2
\end{align}
where we recall that $\lambda_{\max}$ is defined in \eqref{eq:DefLMax}.
\subsubsection{Dynamic Safe sphere \cite{Bonnefoy2015}}
With the dual feasible point defined in (\ref{eq:theta_dynamic}), \eqref{eq:basic_sphere} yields $\mathcal{R} = B(\vt{c},R)$ with
\begin{align} \label{eq:dynamic_sphere}
\text{Dynamic Safe: }  \quad 
\vt{c} =\vt{y}/\lambda, R = \|\vt{\theta}_t - \vt{y}/\lambda \|_2.
\end{align}

\subsubsection{GAP Safe sphere \cite{Fercoq2015}}\label{sssec:GAP_safe_sphere}
Given any primal-dual feasible pair $(\vt{x}_t,\vt{\theta}_t) \in \mathbb{R}^K \!\times\! \Delta_{\mt{A}}$ (say the solution estimations at iteration $t$) and denoting 
\begin{align}\label{eq:duality_gap}
G(\vt{x}_t, \vt{\theta}_t|\mt{A}) := P(\vt{x}_t|\mt{A})  - D(\vt{\theta}_t)
\end{align}
the corresponding duality gap, the GAP Safe sphere is $\mathcal{R} = B(\vt{c},R)$ with
\begin{align} \label{eq:GAP_sphere}
\text{GAP Safe: }  \quad 
\vt{c} = \vt{\theta}_t, R = \frac{1}{\lambda}\sqrt{2 G(\vt{x}_t,\vt{\theta}_t|\mt{A})}.
\end{align}
This region is provably safe for any dual feasible point $\vt{\theta}_t \!\in\! \Delta_{\mt{A}}$  \cite[Theorem 2]{Fercoq2015}. The authors, following \cite{Bonnefoy2015}, propose to use $\vt{\theta}_t$ defined by (\ref{eq:theta_dynamic}).

Note that the cited safe regions reuse quantities computed along the iterations of classical proximal algorithms such as the residual $\vt{\rho}_t$,  the inner products $\mt{A}^{T}\vt{\rho}_t$, or the duality gap (which typically serves as a stopping criterion).

\section{Stable Safe Screening}\label{sec:screening_approx}

When iterating with $\tilde{\mt{A}}$ instead of $\mt{A}$, the computational overhead of accessing inner products $\vt{a}^{T}\vt{c}$ is no longer moderate, hence the screening techniques reviewed in Section \ref{sec:screening} cannot be directly applied in this context. A possibility would of course be to recompute residuals / inner products / duality gaps associated to $\mt{A}$ (instead of $\tilde{\mt{A}}$) to implement the above screening tests, but the whole purpose of re-using quantities already computed with a structured (computationally efficient) matrix $\tilde{\mt{A}}$ would be lost. 

In this section, we propose an alternative solution by deriving screening rules which remain safe w.r.t the original problem \eqref{eq:lasso} {\em even when manipulating an approximate version of the dictionary}. These {\em stable} screening rules only requires some knowledge on the magnitude of certain approximation errors between $\tilde{\mt{A}}$ and $\mt{A}$.

\subsection{Screening a \emph{zone} given a safe region}\label{ssec:error_region}

In conventional screening tests (Definition \ref{def:screening_test}) the dual solution $\vt{\theta}^\star$ is assumed to belong to a known safe region $\mathcal{R}$. This allows to screen (or not) an atom $\vt{a}$ that is perfectly known, (re)using computations of inner products \emph{with this atom}. 

When iterating with $\tilde{\mt{A}}$, efficient screening tests are only entitled to reuse inner products with approximated atoms, $\tilde{\mt{A}}^{T}\vt{c}$. Yet, they should be able to screen certain atoms $\vt{a}$, using the only knowledge that these are ``close'' to the corresponding approximated atoms $\tilde{\vt{a}}$.

One way to capture this knowledge is to consider  $\mathcal{S} \subset \mathbb{R}^{N}$ some neighborhood of $\vt{a}$, assumed to be known. We will call $\mathcal{S}$ a {\em zone}. This gives rise to screening tests for zones.

\begin{definition}[Screening test for a zone] \label{def:screening_test_approx}
Given a safe region $\mathcal{R}$, a screening test 
for the zone $\mathcal{S} \subset \mathbb{R}^{N}$ is given by:
\begin{align}\label{eq:DefZoneTest}
\test{\mathcal{S}}{\mathcal{R}} := \sup_{\vt{a} \in \mathcal{S}} \test{\vt{a}}{\mathcal{R}} = \sup_{\vt{a} \in \mathcal{S}} \sup_{\vt{\theta} \in \mathcal{R}}  |\vt{a}^T \vt{\theta}|
\end{align}
\end{definition}

A sufficient condition for an atom $\vt{a}_{j}$ to be inactive can be expressed as follows: if $\mathcal{R}$ is safe and if $\vt{a}_{j} \in \mathcal{S}$, then 
$$\test{\mathcal{S}}{\mathcal{R}}<1 \implies \vt{x}^\star(j) = 0.$$

\subsection{{\em Stable} screening tests
} \label{ssec:screening_tests_approx}

Given an approximate dictionary $\tilde{\mt{A}}$, error bounds $\epsilon_{j}$ such that $\|\tilde{\vt{a}}_{j}-\vt{a}_{j}\|_{q} \leq \epsilon_{j}$, $1 \leq j \leq K$, can be pre-computed for some choice of $q \in [1,\infty]$, yielding spherical zones $\mathcal{S}_{j} := B_{q}(\tilde{\vt{a}}_{j},\epsilon_{j})$ known to contain the atoms: $\vt{a}_{j} \in \mathcal{S}_{j}$. Using~\eqref{eq:sphere_test},  a spherical zone
\begin{equation}\label{eq:DefBallZone}
\mathcal{S} := B_{q}(\tilde{\vt{a}},\epsilon)
\end{equation}
gives rise to the following estimate: 
\begin{align} \label{eq:sphere_test_approx}
\test{\mathcal{S}}{B_p(\vt{c},R)} \leq |\tilde{\vt{a}}^T \vt{c} |+\epsilon \|\vt{c}\|_{q^*} + R\|\tilde{\vt{a}}\|_{p^*} + C R\epsilon.
\end{align}
where $C = C_{p,q} := 
N^{\left(1-\frac{1}{p} - \frac{1}{q}\right)_{+}}$
and $q^{*} \in [1,\infty]$ is such that $\frac{1}{q}+\frac{1}{q^*} = 1$. See proof in Appendix~\ref{apx:stable_sphere}.

In practice, the norms $a_{j}\!=\!\|\vt{a}_{j}\|_{p^{*}}$ can be pre-computed and stored with negligible overhead, yielding restricted spherical zones  
\begin{align}\label{eq:DefBallNormZone}
\mathcal{S}' := \mathcal{S} \cap \{ \vt{a}: \|\vt{a}\|_{p^{*}} = a\}.
\end{align}
For such zones we get
\begin{align} \label{eq:sphere_test_approx_normalized}
\test{\mathcal{S}'}{B_p(\vt{c},R)} = |\tilde{\vt{a}}^T \vt{c} |+\epsilon \|\vt{c}\|_{q^*} + R a. 
\end{align}
\begin{lemma} \label{lemma:sphere_test_approx}
Consider a dictionary $\mt{A}$ and $\mathcal{R} = B_{p}(\mathbf{c},R)$ a safe region with respect to the dual problem \eqref{eq:lasso_dual}. Let $\tilde{\mt{A}}$ be an approximate dictionary, $1 \leq j \leq K$, and $\mathcal{S}_{j}$ and $\mathcal{S}'_{j}$ defined by~\eqref{eq:DefBallZone}-\eqref{eq:sphere_test_approx_normalized} with $\tilde{\vt{a}} = \tilde{\vt{a}}_{j}$,  $\epsilon \geq \|\tilde{\vt{a}}_{j} - \vt{a}_{j}\|_{q}$, and $a = \|\vt{a}_{j}\|_{p^{*}}$.
\begin{itemize}
\item If $\test{\mathcal{S}_{j}}{B_{p}(\mathbf{c},R)} < 1$ then $\vt{x}_{j}^{\star} = 0$;
\item 
If $\test{\mathcal{S}'_{j}}{B_{p}(\mathbf{c},R)} < 1$ then $\vt{x}_{j}^{\star} = 0$.
\end{itemize}
\end{lemma}

\begin{remark}
In \cite{Herzet2018}, Definition \ref{def:screening_test_approx} is used in a different context. No approximate atoms are considered and the idea is rather to simultaneously test multiple atoms which eventually lie within a same zone $\mathcal{S}$, reducing the total number of tests performed.
\end{remark}

\subsection{Building a safe region using an approximate dictionary} \label{ssec:approx_safe_region}

As in classical screening, we have all ingredients to screen provided we can build a safe region using moderate computational overhead. The object of this section is precisely to adapt the constructions of safe regions from Section~\ref{ssec:safe_regions}, which depend on $\mt{A}$, to define new safe regions reusing computations done during the iterations {\em with $\tilde{\mt{A}}$ instead of $\mt{A}$}. 

\begin{remark}
Let us emphasize that a safe region is a region $\mathcal{R}$ containing the dual solution $\vt{\theta}^\star$ of the \emph{original} dual problem \eqref{eq:lasso_dual} and \emph{not necessarily} the dual solution of its approximate version with $\Delta_{\tilde{\mt{A}}}$ instead of $\Delta_\mt{A}$. Indeed, although $\mt{A}$ is approximated to speed up computations,  variable elimination needs to be guaranteed with respect to the \emph{original} problem~\eqref{eq:lasso}.
\end{remark}

The safe regions from Section~\ref{ssec:safe_regions} are built by determining a feasible dual point $\vt{\theta}_t \!\in\! \Delta_{\mt{A}}$.
Given an arbitrary $\vt{z}\in \mathbb{R}^N$, the function $\Theta(\vt{z} | \mt{A})$ from \eqref{eq:conditional_dual_scaling} would provide such a feasible point, however it cannot be used as its computation requires multiplication by $\mt{A}^{T}$.
A naive alternative could be to compute $\Theta(\vt{z}|\tilde{\mt{A}})$, however while this always belongs to $\Delta_{\tilde{\mt{A}}}$ (the feasible set for $\tilde{\mt{A}}$) it does not necessarily belong to the desired feasible set $\Delta_{\mt{A}}$. This can be fixed using a modified dual scaling approach that we propose to call {\em stable} dual scaling.

Considering error bounds $\vt{\epsilon} = (\epsilon_{j})_{j=1}^{K} \in \mathbb{R}_{+}^{K}$ we define

\begin{align} \label{eq:conditional_dual_scaling_modified}
\Theta'(\vt{z}|\tilde{\mt{A}},\vt{\epsilon}) := \frac{\vt{z}}{\max\left(1, \max_{1 \leq j \leq K} ( |\tilde{\vt{a}}^T_j \vt{z}| + \epsilon_{j} \|\vt{z}\|_{q^*} ) \right)}
\end{align}
For $\vt{\epsilon}=\vt{0}$ we recover $\Theta(\vt{z}|\mt{A}) = \Theta'(\vt{z}|\mt{A},\vt{0})$.
\begin{lemma} \label{lemma:theta_dynamic_approx}
Assume  that $\vt{a}_{j} \in \mathcal{S}_{j} = B_{q}(\tilde{\vt{a}}_{j},\epsilon_{j})$ for $1 \leq j \leq K$. Then,
for any $\vt{z}\in \mathbb{R}^N$, $\Theta'(\vt{z}|\tilde{\mt{A}},\vt{\epsilon}) \in \Delta_{\mt{A}}  \cap \Delta_{\tilde{\mt{A}}}$ is a feasible point w.r.t. both the original dual problem \eqref{eq:lasso_dual} and its modified version with $\tilde{\mt{A}}$ instead of $\mt{A}$.
\end{lemma}
The proof is in Appendix~\ref{apx:2}. Analogously to \eqref{eq:theta_dynamic}, we can now define a dual feasible point proportional 
to the residual $\tilde{\vt{\rho}}_t \!:=\! \vt{y} \!-\! \tilde{\mt{A}}\vt{x}_t$ at iteration $t$,
$\vt{\theta}'_t = \Theta'\left(\proj{\dfrac{\vt{y}}{\lambda}}{\tilde{\vt{\rho}}_t}  ~\vert ~ \tilde{\mt{A}}, \vt{\epsilon} \right)$,
that is to say more explicitly
\begin{align} \label{eq:theta_dynamic_approx}
&\vt{\theta}'_t 
= \left[\dfrac{\vt{y}^T \tilde{\vt{\rho}}_t}{\lambda \|\tilde{\vt{\rho}}_t\|_2^2}\right]_{-\alpha_t'}^{~ \alpha'_t} \tilde{\vt{\rho}}_t, \nonumber \\
& \text{with } 
\alpha'_t = \left( \max_{1 \leq j \leq K} \left( |\tilde{\vt{a}}^T_j \tilde{\vt{\rho}}_t| + \epsilon_{j} \|\tilde{\vt{\rho}}_t\|_{q^*} \right) \right)^{-1}
\end{align}

\begin{table*}
\renewcommand{\arraystretch}{2}
\centering
	\begin{tabular}{|c||c|c||c|}
	\hline
	\textbf{Sphere test} & \textbf{Center} 			& \textbf{Radius}  
						& \textbf{Test expression} - $\test{\vt{a}}{B(\vt{c},R)}$ or  $\test{
						\mathcal{S}'}{B(\vt{c}',R')}$ \\  \hline 	\hline
	Static Safe 		& $\vt{y}/\lambda$			& $|1/\lambda_{\max}-1/\lambda| \cdot \|\vt{y}\|_{2}$  
						& $|\vt{a}^T \vt{y}|/\lambda + |1/\lambda_{\max}-1/\lambda| \cdot \|\vt{y}\|_{2} \cdot \|\vt{a}\|_{2}$ \\ \hline
	Stable Static Safe 	& $\vt{y}/\lambda$ 		& $|1/\lambda'_{\max}-1/\lambda| \cdot \|\vt{y}\|_{2}$  
						& $|\tilde{\vt{a}}^T \vt{y}|/\lambda + \epsilon \|\vt{y}\|_{q^*}/\lambda + |1/\lambda'_{\max}-1/\lambda| \cdot \|\vt{y}\|_{2} \cdot \|\vt{a}\|_{2}$ \\ \hline \hline			
	Dynamic Safe 		& $\vt{y}/\lambda$			& $\|\vt{\theta}_t - \vt{y}/\lambda \|_2$  
						& $|\vt{a}^T \vt{y}|/\lambda + \|\vt{\theta}_t - \vt{y}/\lambda \|_2 \cdot \|\vt{a}\|_2$ \\ \hline
	\multirow{2}{*}{Stable Dynamic Safe} & \multirow{2}{*}{$\vt{y}/\lambda$} 		& \multirow{2}{*}{$\|\vt{\theta}'_t - \vt{y}/\lambda \|_2$}						
						&  $|\vt{a}^T \vt{y}|/\lambda + \|\vt{\theta}'_t - \vt{y}/\lambda \|_2 \cdot \|\vt{a}\|_{2}$, \hfill if $\vt{a}^T\vt{y}$ is given\footnotemark[2]\\  \cline{4-4}
					& & & $|\tilde{\vt{a}}^T \vt{y}|/\lambda + \epsilon \|\vt{y}\|_{q^*}/\lambda + \|\vt{\theta}'_t - \vt{y}/\lambda \|_2 \cdot \|\vt{a}\|_{2}$, \quad  \hfill otherwise \\  \hline					\hline
	GAP Safe 			& $\vt{\theta}_t$ 			& $\frac{1}{\lambda}\sqrt{2G(\vt{x}_t,\vt{\theta}_t|\mt{A})}$  
						& $|\vt{a}^T \vt{\theta}_t |+ \frac{1}{\lambda}\sqrt{2G(\vt{x}_t,\vt{\theta}_t|\mt{A})} \cdot \|\vt{a}\|_{2}$ \\ \hline
	Stable GAP Safe 	& $\vt{\theta}'_t$ 	& $\frac{1}{\lambda}\sqrt{2 G(\vt{x}_t,\vt{\theta}'_t|\tilde{\mt{A}}) + 2\delta(\vt{x}_t) }$ 
						& $|\tilde{\vt{a}}^T \vt{\theta}'_t |+\epsilon \|\vt{\theta}'_t\|_{q^*} + \frac{1}{\lambda}\sqrt{2 G(\vt{x}_t,\vt{\theta}'_t|\tilde{\mt{A}}) + 2\delta(\vt{x}_t) } \cdot \|\vt{a}\|_{2}$ \\ \hline
	\end{tabular}
\caption{Sphere center $\vt{c}$ and radius $R$ for Static Safe, Dynamic Safe and GAP Safe ($\vt{c}'$ and $R'$ for the stable versions). \\
See definitions of $\lambda_{\max}$ in \eqref{eq:DefLMax}, of $\lambda'_{\max}$ in \eqref{eq:DefStableLMax}, of $\vt{\theta}_{t}$ in \eqref{eq:theta_dynamic}, of $\vt{\theta}'_{t}$ in \eqref{eq:theta_dynamic_approx},  and of $\delta(\vt{x}_t)$ in \eqref{eq:DefDelta} with $\vt{x} = \vt{x}_{t}$. \\
Test expressions for $\test{\vt{a}}{B(\vt{c},R)}$ -- see \eqref{eq:sphere_test} -- and $\test{\mathcal{S}'}{B(\vt{c}',R')}$ -- see \eqref{eq:sphere_test_approx_normalized} where $\mathcal{S}'$ is defined in \eqref{eq:DefBallNormZone} with $a = \|\vt{a}\|_{2}$.}
\label{tab:screening_rules}
\end{table*}

These principles yield new {\em stable} safe $\ell_{2}$-spherical regions.
\subsubsection{Stable Static Safe sphere}\
The static safe sphere \eqref{eq:static_sphere} depends on $\lambda_{\max} = \max_{j} |\vt{a}_{j}^{T}\vt{y}|$. To perform only computations with $\tilde{\mt{A}}$, we can reuse \eqref{eq:static_sphere} with 
\begin{equation}\label{eq:DefStableLMax}
\lambda'_{\max} := \max_{j} \left(|\tilde{\vt{a}}_{j}^{T}\vt{y}|+\epsilon_{j} \|\vt{y}\|_{q^{*}}\right) \geq \lambda_{\max},
\end{equation}
 leading to $\mathcal{R} = B(\vt{c}',R')$ with
\begin{align}\label{eq:stable_static_sphere}
\text{Stable Static Safe: }  \quad 
\vt{c}'=\vt{y}/\lambda, R'=|\tfrac{1}{\lambda'_{\max}} - \tfrac{1}{\lambda}| \cdot \|\vt{y}\|_2.
\end{align}
\subsubsection{Stable Dynamic Safe sphere}\
We adapt the Dynamic Safe sphere in \eqref{eq:dynamic_sphere} to our approximate setting by replacing the dual feasible point $\vt{\theta}_t$ \eqref{eq:theta_dynamic} by $\vt{\theta}'_t$ \eqref{eq:theta_dynamic_approx}, leading to $\mathcal{R} = B(\vt{c}',R')$ with
\begin{align} \label{eq:dynamic_Safe_sphere_approx}
\text{Stable Dynamic Safe:} \quad 
\vt{c}' = \vt{y}/\lambda, R' = \|\vt{\theta}'_t - \vt{y}/\lambda \|_2.
\end{align}
The fact that $\vt{\theta}'_t \in \Delta_{\mt{A}}$ directly implies that this sphere is safe by definition of the basic $\ell_2$-spherical bound \eqref{eq:basic_sphere}.

\

\subsubsection{Stable GAP Safe sphere}\
To build a safe sphere $B(\vt{c}',R')$ with $\vt{c}' = \vt{\theta}'_t$ \eqref{eq:theta_dynamic_approx} instead of $\vt{c}=\vt{\theta}_t$ \eqref{eq:theta_dynamic} we need to determine a safe radius $R'$
depending only on $\tilde{\mt{A}}$. For this reason, instead of using the (inaccessible) duality gap $G(\vt{x}_t,\vt{\theta}'_t|\mt{A})=P(\vt{x}_t|\mt{A}) - D(\vt{\theta}'_t)$ we will use $G(\vt{x}_t,\vt{\theta}'_t|\tilde{\mt{A}}) = P(\vt{x}_t|\tilde{\mt{A}}) - D(\vt{\theta}'_t)$ and add a security margin ensuring that $B(\vt{\theta}'_t, R')$ is safe. Recall the standard notation $\|\mt{M}\|_{p \to q} := \sup_{\|u\|_{p}\leq 1}\|\mt{M}\vt{u}\|_{q}$.

\begin{comment}
We reason as follows:
\begin{itemize}
\item $\vt{\theta}'_t$ is feasible with respect to $\mt{A}$, i.e. $\vt{\theta}'_t \in \Delta_{\mt{A}}$.
\item A GAP safe sphere \eqref{eq:GAP_sphere} would correspond to a safe radius $R \!=\! \frac{1}{\lambda}\sqrt{2 G(\vt{x}_t,\vt{\theta}'_t|\mt{A})}$  $= \frac{1}{\lambda}\sqrt{2 ( P(\vt{x}_t|\mt{A}) - D(\vt{\theta}'_t) )}$.
This cannot be calculated since $P(\vt{x}_t|\mt{A})$ depends on $\mt{A}$.
\item A computable quantity is $G(\vt{x}_t,\vt{\theta}'_t|\tilde{\mt{A}}) = P(\vt{x}_t|\tilde{\mt{A}}) - D(\vt{\theta}'_t).$
As $\vt{\theta}'_t$ is also feasible with respect to $\tilde{\mt{A}}$, i.e. $\vt{\theta}'_t \in \Delta_{\tilde{\mt{A}}}$ (cf Lemma~\ref{lemma:theta_dynamic_approx}), $G(\vt{x}_t,\vt{\theta}'_t|\tilde{\mt{A}}) \geq 0$ is indeed a duality gap associated to the \emph{approximate primal} $P(\vt{x}_t|\tilde{\mt{A}})$  and its dual. However the sphere $B(\vt{\theta}'_{t},R'')$ with $R'' \!=\! \frac{1}{\lambda}\sqrt{2 G(\vt{x}_t,\vt{\theta}'_t|\tilde{\mt{A}})}$ is not a priori a safe region  wrt to the original primal $P(\vt{x}_t|\mt{A})$.
\end{itemize}
Therefore, we add a computable security margin to $R''$ in order to get a radius $R' \geq R$, ensuring that $B(\vt{\theta}'_t, R')$ is safe. Recall the standard notation $\|\mt{M}\|_{p \to q} := \sup_{\|u\|_{p}\leq 1}\|\mt{M}\vt{u}\|_{q}$.
\end{comment}

\begin{theorem}[Stable GAP Safe sphere] \label{thm:GAP_approx}
Consider 
\[
\mathcal{E} \geq \|\mt{A}-\tilde{\mt{A}}\|_{r \rightarrow 2}
\]
where $1 \leq r  \leq \infty$. 
For any\footnote{In this theorem, $\theta \in \Delta_{\mt{A}}$ can be arbitrary. In particular it is not necessarily expressed as in \eqref{eq:theta_dynamic_approx}, hence it does not need to belong to $\Delta_{\tilde{\mt{A}}}$.} $(\vt{x},\vt{\theta}) \in \mathbb{R}^K \!\times\! \Delta_{\mt{A}}$, we have
\begin{align}\label{eq:GAP_sphere_extended}
& \vt{\theta}^\star \in B \left(\vt{c}' = \vt{\theta}, R'= \frac{1}{\lambda}\sqrt{2 G(\vt{x},\vt{\theta}|\tilde{\mt{A}}) + 2\delta(\vt{x}) } \right) \\
& \text{with } \qquad
 \delta(\vt{x}) := \|\vt{y}-\tilde{\mt{A}}\vt{x}\|_2 \mathcal{E}\|\vt{x}\|_r + \frac{1}{2}  \mathcal{E}^2\|\vt{x}\|_r^2\label{eq:DefDelta}
\end{align}
In other words $B(\vt{\theta},R')$ is safe.
\end{theorem}
The proof was given in \cite{F.Dantas2018}.
By applying Theorem \eqref{thm:GAP_approx} to $\vt{x}_t$ and $\vt{\theta}'_t$ in \eqref{eq:theta_dynamic_approx} we obtain $\mathcal{R} = B(\vt{c}',R')$ with
\begin{align} \label{eq:dynamic_Safe_sphere_approx}
\text{Stable GAP Safe:} \quad 
\vt{c}' \!=\! \vt{\theta}'_t, R' \!=\!
\frac{1}{\lambda}\sqrt{2 G(\vt{x}_t,\vt{\theta}'_t|\tilde{\mt{A}}) + 2\delta(\vt{x_t})}.
\end{align}

\

The first columns of Table \ref{tab:screening_rules} summarize the obtained safe regions alongside with their previously existing analogue.

Combining stable safe sphere regions with stable safe screening tests (cf Section \ref{ssec:screening_tests_approx}) yields screening rules which remain safe despite manipulating an approximate dictionary (last column of Table \ref{tab:screening_rules}).
As before, these new rules also reuse the calculations performed by optimization algorithm with the approximate dictionary $\tilde{\mt{A}}$, making them suitable in practice. 

\begin{comment}
For stable regions corresponding to the first four lines of Table~\ref{tab:screening_rules}, the center $\vt{c} = \vt{y}/\lambda$ is independent of the iteration count $t$. In a scenario where $\mt{A}^T\vt{c} = \mt{A}^{T}\vt{y}/\lambda$ is calculated once, this implies that:
\begin{itemize}
\item the stable static rule loses any interest as the sharper plain static rule can be computed ($\lambda_{\max} = \|\mt{A}^{T}\vt{c}\|_{\infty}$);
\item the only difference between the stable and plain Dynamic Safe test comes from the radius $R'$ of the stable safe region which uses $\vt{\theta}'_t$ instead of  $\vt{\theta}_t$ (see Table \ref{tab:screening_rules}, row four).  
\end{itemize}
The resulting tests are given in the last columns of Table \ref{tab:screening_rules}.
\end{comment}

\section{A Fast Algorithm for $\ell_{1}$ regularization} 
\label{sec:algorithm}

\stepcounter{footnote}\footnotetext{Since the center $\vt{c} = \vt{y}/\lambda$ is independent of the iteration count $t$, the resulting test can be simplified provided that $\mt{A}^{T}\vt{y}$ can be pre-computed.}

We now have the tools to combine stable safe screening and the use of fast structured dictionaries. The proposed Algorithm~\ref{alg:ISTA_dynamic_approx} consists in incorporating a stable dynamic screening rule to iterations of a proximal algorithm.
Besides the vector $\vt{y}$ and the regularization parameter $\lambda$, the input of the algorithm consists in a sequence $\{\tilde{\mt{A}}^{i}\}_{i=0}^{I}$ of approximate dictionaries with $\tilde{\mt{A}}^{I}=\mt{A}$, and a corresponding sequence of error bounds $\{\vt{\epsilon}^{i}\}_{i=0}^{I}$, where $\vt{\epsilon}^{i}=(\epsilon^i_j)_{j=1}^K \in \mathbb{R}_{+}^{K}$ contains $\ell_{2}$ error bounds on the atoms of the $i$-th dictionary $\tilde{\mt{A}}^{i}$, and of course $\vt{\epsilon}^{I} = \vt{0}$.

In general terms, the proposed strategy consists in gradually switching to more accurate dictionary approximations 
as the optimization algorithm approaches the solution.
Note that any iterative optimization technique can be used by replacing the generic update function $p(\vt{x}_{t},\mt{A},\vt{\alpha}_t)$ by the corresponding update step, for instance equation \eqref{eq:ISTA_update} for ISTA. A possible variation consists in performing the screening at regular intervals instead of every iteration.

\begin{algorithm}
\caption{$\vt{x}^\star = \mathtt{FastL1}
(\{\tilde{\mt{A}}^i\}_{i=0}^{I},\{\vt{\epsilon}^i\}_{i=0}^{I},\vt{y},\lambda,\Gamma)$
} \label{alg:ISTA_dynamic_approx}
\begin{algorithmic} [1]
\State \textbf{Initialize:} $t=0$, $\vt{x}_0 = \vt{0}$, $i=0$, $\mathcal{A}_{0}=\{1,\dots,K\}$,\\ $\vt{\alpha}_{0}$ according to the considered proximal algorithm
	\While {stopping criterion not met}
		\State \textit{----- Restrict to preserved set -----}
		\State $\tilde{\mt{A}} \gets \tilde{\mt{A}}^{i}_{[\mathcal{A}_{t}]}$, $\quad \vt{x}_{t} \gets (\vt{x}_{t})_{[\mathcal{A}_{t}]}$ 
		\State \textit{----- Update preserved coefficients -----}
		\State $\{\mt{x}_{t+1},\vt{\alpha}_{t+1}\} \gets p(\mt{x}_{t},\tilde{\mt{A}},\vt{\alpha}_{t})$
		\State \textit{----- Dynamic Screening -----}
		\State Set $\vt{\theta}'_t$ using (\ref{eq:theta_dynamic_approx}) and $\vt{c}'$, $R'$ according to Table~\ref{tab:screening_rules}
		\State $\mathcal{A}_{t+1} \gets \{j \!\in\! \mathcal{A}_{t} : \test{\mathcal{S}'_{j}}{B(\vt{c}',R')} \geq 1 \}$
		\State $t \gets t+1$
		\State \textit{----- Switching criterion -----}
		\State Compute $\gamma_{t}$ (cf. \eqref{eq:DefRatioSwitch}) and $K_{t}$ (cf. \eqref{eq:DefCardinalEstimated})
		\State $i = \mathtt{SwitchDictionary}(i,I,\gamma_{t},\Gamma,K_{t})$
	\EndWhile
\end{algorithmic}
\end{algorithm}

To fully describe this algorithm we need to specify the stopping and switching criteria, which involve a parameter $\Gamma$. 
In the following subsections,
we present a switching criterion which assumes the approximations $\{\tilde{\mt{A}}\}_{i=0}^I$ to be arranged in increasing order of precision.

A classical stopping criterion is to stop when the duality gap falls below a threshold, recalling that strong duality holds for problem \eqref{eq:lasso}, i.e. $G(\vt{x}^\star, \vt{\theta}^\star|\mt{A}) \!=\! 0$~\cite{Osborne2000}. The simplest way of computing the duality gap is to wait until the algorithm reaches $\tilde{\mt{A}} = \tilde{\mt{A}}^{I} = \mt{A}$. Alternatively one could upper bound the duality gap using approximate dictionaries, this is left to future work. 

A switching criterion has two main motivations:
\begin{itemize}
\item {\bf Convergence:} to avoid diverging from the solution of the original problem \eqref{eq:lasso}. The higher the error bounds $\vt{\epsilon}^{i}$, the higher the risk of moving away from the true solution, therefore the need to switch earlier in such cases.
\item {\bf Speed:} once screened enough, the original dictionary becomes faster to apply than the approximate ones.
This is crucial at high regularization regimes ($\lambda/\lambda_{\max} \!\approx\! 1$) in which screening typically occurs very quickly. 
\end{itemize}

\subsection{Switching: convergence criterion}\label{ssec:switching_convergence}\

In \cite{F.Dantas2017,F.Dantas2018}, we proposed a empirically-tuned heuristic convergence criterion. In the following, we derive a more theoretically grounded criterion. 

Given a solution estimate $\vt{x}_t$ at iteration $t$, a natural convergence measure for the problem $\mathcal{L}(\lambda,\mt{A},\vt{y})$ is the duality gap $G(\vt{x}_t, \vt{\theta}_t|\mt{A})$~\eqref{eq:duality_gap} 
with $\vt{\theta}_t$ calculated as in \eqref{eq:theta_dynamic}. However, when manipulating approximate dictionaries,
this quantity is not accessible without unwanted additional computations.

As a surrogate, we propose to use a ratio
\begin{align}\label{eq:DefRatioSwitch}
\gamma_t = \frac{G(\vt{x}_t, \tilde{\vt{\theta}}_t|\tilde{\mt{A}})}{G(\vt{x}_t, \vt{\theta}'_t|\tilde{\mt{A}})}
\end{align}
between two computable gaps arising from two different dual points $\vt{\theta}'_t$ and $\tilde{\vt{\theta}}_t$ that are respectively feasible for problems $\mathcal{L}(\lambda,\mt{A},\vt{y})$ and $\mathcal{L}(\lambda,\tilde{\mt{A}},\vt{y})$:
\begin{align}
\label{item:gap_prime} 
\vt{\theta}'_t \in \Delta_{\mt{A}} \cap \Delta_{\tilde{\mt{A}}} &\longrightarrow G(\vt{x}_t, \vt{\theta}'_t|\tilde{\mt{A}}) = P(\vt{x}_t|\tilde{\mt{A}})  - D(\vt{\theta}'_t)\\
\label{item:gap_tilde} 
\tilde{\vt{\theta}}_t \in \Delta_{\tilde{\mt{A}}} &\longrightarrow G(\vt{x}_t, \tilde{\vt{\theta}}_t|\tilde{\mt{A}}) = P(\vt{x}_t|\tilde{\mt{A}})  - D(\tilde{\vt{\theta}}_t)
\end{align}
where $\vt{\theta}'_t \in \Delta_{\mt{A}} \cap \Delta_{\tilde{\mt{A}}}$ has been previously calculated for the stable screening  with \eqref{eq:theta_dynamic_approx}, while $\tilde{\vt{\theta}}_t \in \Delta_{\tilde{\mt{A}}}$ is the conventional dual point \eqref{eq:theta_dynamic} but calculated with $\tilde{\mt{A}}$.
Then again, apart from minor extra memory requirements, the computation of the quantities in \eqref{item:gap_tilde} reuses most of the calculations%
\footnote{Only $N$ extra products for computing $\|\tilde{\vt{\theta}}\|$ and $|\mathcal{A}_t|$ comparisons for $\|\tilde{\mt{A}}^T\vt{\rho}_t\|_\infty$ are required.}
in \eqref{item:gap_prime}.

\begin{figure}[t]
	\centering
    \includegraphics[width=\linewidth,trim={0 0.2cm 0cm 0cm},clip]{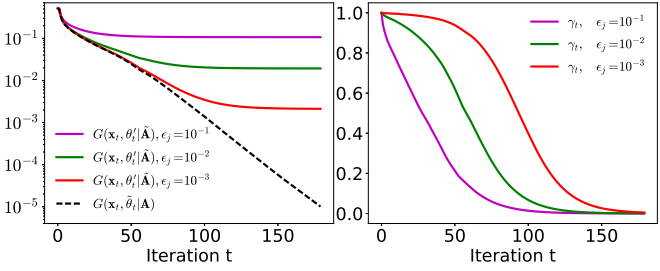}
    \caption{Typical behaviour of duality gaps (left) and ratio $\gamma_t$ (right) as a function of the iteration number and error bounds. \label{fig:duality_gap_saturation}}
\end{figure}

If $\tilde{\mt{A}}=\tilde{\mt{A}}^{i}$ was kept until convergence, we would have
\begin{align}
\tilde{\vt{x}}_t \rightarrow \tilde{\vt{x}}^{\star}
& & \text{and} & &
\tilde{\vt{\theta}}_t \rightarrow \tilde{\vt{\theta}}^{\star}.
\end{align}
with $\tilde{\vt{x}}^{\star}$ a solution of $\mathcal{L}(\lambda,\tilde{\mt{A}},\vt{y})$ and $\tilde{\vt{\theta}}^{\star}$ a solution of the corresponding dual problem. Then,  as illustrated in Fig. \ref{fig:duality_gap_saturation} (left), the second quantity $G(\vt{x}_t, \tilde{\vt{\theta}}_t|\tilde{\mt{A}})$ --which is precisely the duality gap w.r.t. the problem $\mathcal{L}(\lambda,\tilde{\mt{A}},\vt{y})$-- would tend to zero, while the first quantity $G(\vt{x}_t, \vt{\theta}'_t|\tilde{\mt{A}})$ would typically saturate at a nonzero value since $\vt{\theta}'_t \nrightarrow \tilde{\vt{\theta}}^{\star}$. 
As a result, the ratio $\gamma_t$ would shrink with the iterations as illustrated in Fig. \ref{fig:duality_gap_saturation} (right).

With that in mind, we propose to switch dictionaries when
\begin{equation}
\label{eq:saturation_threshold}
\gamma_{t}  \leq \Gamma
\end{equation}
where the higher $\Gamma$, the greater sensibility to the gap saturation.

\subsection{Speed criterion}\label{ssec:switching_complexity}\

In terms of speed, the general idea is to exploit tradeoffs between approximation and computational efficiency.

Denoting $\mathcal{C}(\mt{A}, \mathcal{A})$ the computational cost of multiplying matrix $\mt{A}_{[\mathcal{A}]}$ by a vector, it is reasonable to assume that $\mathcal{C}(\mt{A}, \mathcal{A}) \leq 
\mathcal{C}(\mt{A}, \mathcal{A}')$ if $\mathcal{A} \subset \mathcal{A}'$.
As a result, given the nested structure of the sequence of preserved sets $\mathcal{A}_{t}$, the sequence $\mathcal{C}(\mt{A}, \mathcal{A}_{t})$, $t \geq 0$ is non-increasing.

Consider two approximate dictionaries $\tilde{\mt{A}}^{i}$ and $\tilde{\mt{A}}^{j}$ and assume the initial ``unscreened'' computational costs satisfy $\mathcal{C}(\tilde{\mt{A}}^{i}, \mathcal{A}_{0}) < \mathcal{C}(\tilde{\mt{A}}^{j}, \mathcal{A}_{0})$. Denote $\mt{E}^{i}$, $\mt{E}^{j}$ the corresponding approximation errors (cf \eqref{eq:approx_dict}), and assume that $\mt{E}^{i}$ is ``larger'' than $\mt{E}^{j}$ (let us denote this $\mt{E}^{i} \succ \mt{E}^{j}$) in the sense that for each column the approximation error is larger. An irrevocable switching point would be an iteration $T$ in which the coarser approximation $\tilde{\mt{A}}^{i}$ becomes more complex than $\tilde{\mt{A}}^{j}$ 
\begin{align} \label{eq:switching_ratio_criterion_general}
\forall t > T, \quad \mathcal{C}(\tilde{\mt{A}}^{i}, \mathcal{A}_{t}) \geq \mathcal{C}(\tilde{\mt{A}}^{j}, \mathcal{A}_{t}).
\end{align}
Note that this only happens if
$\mathcal{C}(\tilde{\mt{A}}^{i}, \mathcal{A}_{t})$ decreases faster than $\mathcal{C}(\tilde{\mt{A}}^{j}, \mathcal{A}_{t})$  as screening progresses (i.e. as $|\mathcal{A}_{t}|$ gets smaller).

To use the general idea just described, one needs to specify a complexity model $\mathcal{C}(\mt{A}, \mathcal{A}_{t})$ depending on the structure of the corresponding matrix. 
For a generic unstructured dictionary matrix $\mt{A} \in \mathbb{R}^{N \times K}$, a simple model is:
\begin{align} \label{eq:complexity_model}
\mathcal{C}(\mt{A}, \mathcal{A}_{t}) = N |\mathcal{A}_{t}|
\end{align}

To quantify the complexity reduction on matrix-vector products entailed by an approximate dictionary
we will use the concept of \emph{Relative Complexity} (RC) \cite{Magoarou2016}, such that
\begin{align} \label{eq:complexity_model_approx}
\mathcal{C}(\tilde{\mt{A}}, \mathcal{A}_{0}) = \text{RC}(\tilde{\mt{A}}) \!\times\! NK.
\end{align}
In a worst-case scenario, screening does not further reduce the cost and 
$\mathcal{C}(\tilde{\mt{A}}, \mathcal{A}_{t}) = \text{RC}(\tilde{\mt{A}}) \!\times\! NK$.
In more optimistic scenarios, some fast approximate dictionary structures might still benefit from speedups upon column removal.

Consider approximate dictionaries $\{\tilde{\mt{A}}^{0}, \dots, \tilde{\mt{A}}^{I} \}$, with $\tilde{\mt{A}}^{I}= \mt{A}$,
of increasing precision and decreasing complexity
\begin{align}
\mt{E}^{1} \succ \ldots \mt{E}^{I}\!=\!\mt{0} & ~~\text{and}~~
\text{RC}(\tilde{\mt{A}}^{0}) < \ldots < \text{RC}(\tilde{\mt{A}}^{I}) \!=\! 1.
\end{align}
The above observations suggest to switch from $\tilde{\mt{A}}^{i}$ to $\mt{A}$ if
$\text{RC}(\tilde{\mt{A}}^{i}) \!\times\! NK 
= \mathcal{C}(\tilde{\mt{A}}^{i}, \mathcal{A}_{t})
\geq \mathcal{C}(\mt{A}, \mathcal{A}_{t}) = N |\mathcal{A}_{t}|$, giving
\begin{align} \label{eq:switching_ratio_criterion}
|\mathcal{A}_t| \leq \text{RC}(\tilde{\mt{A}}^{i}) ~K.
\end{align}
Also, as a consequence of complexity model \eqref{eq:complexity_model_approx} adopted for the approximate dictionaries $\tilde{\mt{A}}$, this criterion never causes a switching from $\tilde{\mt{A}}^{i}$  to $\tilde{\mt{A}}^{j}$ with $i<j \neq I$.

\subsection{Heuristic look-ahead speed criterion}

The preserved set size $|\mathcal{A}_{t}|$ used in \eqref{eq:switching_ratio_criterion} should ideally be the one associated to the dictionary {\em after} switching ($\mt{A}$ in this case), while in practice only the screening rate associated to the {\em current} approximation $\tilde{\mt{A}}^i$ is available. 
This motivates the introduction of a heuristic (that preserves the safety of screening) to anticipate the potentially smaller size of the preserved set $|\mathcal{A}_{t}|$ as soon as we switch to a more precise approximation (or to the original dictionary).

\begin{figure}
	\centering
    \includegraphics[width=\linewidth,trim={0 0.3cm 0cm 0cm},clip]{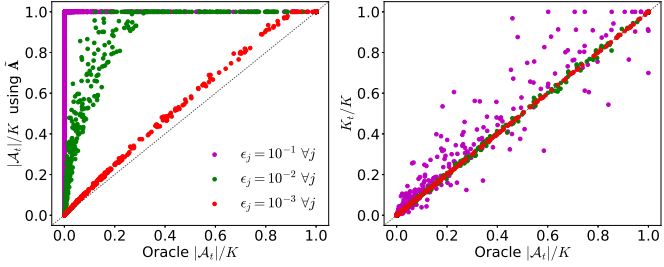}
    \caption{Number of preserved atoms $|\mathcal{A}|$ using oracle conventional screening (x-axis) {\em vs} stable screening (y-axis, left plot), or {\em vs} proposed heuristic (y-axis, right plot). \label{fig:scatterplot_nb_preserved_atoms}.}
\end{figure}

Indeed, stable screening generally leads to less atom eliminations due to the extra security margins (and more so for higher approximation errors). This is illustrated in Figure \ref{fig:scatterplot_nb_preserved_atoms} (left) with a scatterplot in which the x-axis correspond to the ``oracle'' number of preserved atoms $|\mathcal{A}|$ obtained with a conventional Dynamic Safe screening mechanism having access to $\mt{A}$. This is the quantity we are interested in estimating for the speed criterion \eqref{eq:switching_ratio_criterion}. 
The y-axis corresponds to the number of preserved atoms obtained by the corresponding stable test for a given approximation $\tilde{\mt{A}}$. Each plotted point compares the two mentioned screening ratios at a given iteration. The procedure was repeated a large number of times in order to give a full picture of how the two quantities correlate to one another.
The main observation is that all points lie above the identity line which means that the stable test always  overestimates (sometimes significantly) the value that $|\mathcal{A}_{t}|$ will reach after switching. This phenomenon is intensified for higher approximation errors. 

As a heuristic, we propose to estimate $|\mathcal{A}_t|$ using the \emph{conventional} test 
on the approximate atoms, i.e. $\test{\tilde{\vt{a}}_j}{\mathcal{R}}$. 

\begin{equation}\label{eq:DefCardinalEstimated}
K_{t} =  |\{j \in \{1,\dots,K\} : \test{\tilde{\vt{a}}_j}{\mathcal{R}} \geq 1 \}|
\end{equation}
As shown in Fig. \ref{fig:scatterplot_nb_preserved_atoms} (right graph), this correlates much better with the oracle value of $|\mathcal{A}_{t}|$, since it doesn't have the additional security margins used in the stable test. Even though the test used to compute $K_{t}$ is unsafe with respect to the original problem \eqref{eq:lasso}, it has no impact on the safety of screening itself, which is still performed with the stable test.
Moreover, the additional test is virtually costless since the same calculations of the first test are used.

\subsection{Summary and example}

The resulting switching strategy, shown in Algorithm \ref{alg:SwitchDictionary}, plays two roles:  
\begin{enumerate}[label=(\Roman*)]
\item \label{item:role1} Decide when to stop using the current dictionary.
\item \label{item:role2} Choose the next dictionary.
\end{enumerate}

\begin{algorithm}
\caption{$j = \mathtt{SwitchDictionary}(i,I,\gamma_{t},\Gamma,K_{t})$} \label{alg:SwitchDictionary}
\begin{algorithmic} [1]
\If {$K_t \leq \text{RC}(\tilde{\mt{A}}^{i}) ~K$} \Comment{Speed criterion}
	\State $j \gets I$
\ElsIf {$\gamma_t \leq \Gamma$} \Comment{Convergence criterion}
	\State $j \gets i+1$
\Else \State $j \gets i$
\EndIf
\end{algorithmic}
\end{algorithm}

The speed criterion (section \ref{ssec:switching_complexity}) triggers a switching directly to the original dictionary $\tilde{\mt{A}}^I = \mt{A}$. We suppose the Relative Complexity associated to the i-th approximation $\text{RC}(\tilde{\mt{A}}^{i})$ (cf. \eqref{eq:complexity_model_approx}) to be known and stored in memory.
The convergence criterion (section \ref{ssec:switching_convergence}) switches to the next available approximation $\tilde{\mt{A}}^{i+1}$. 
Note that if $\mt{A}$ is already adopted, i.e. input $i=I$, then $\mt{A}$  is kept.

\ 

\begin{figure}
	\centering
	\begin{subfigure}[t]{0.5\linewidth}
        \centering
        \includegraphics[width=\linewidth,trim={0 0.5cm 0 0.3cm}]{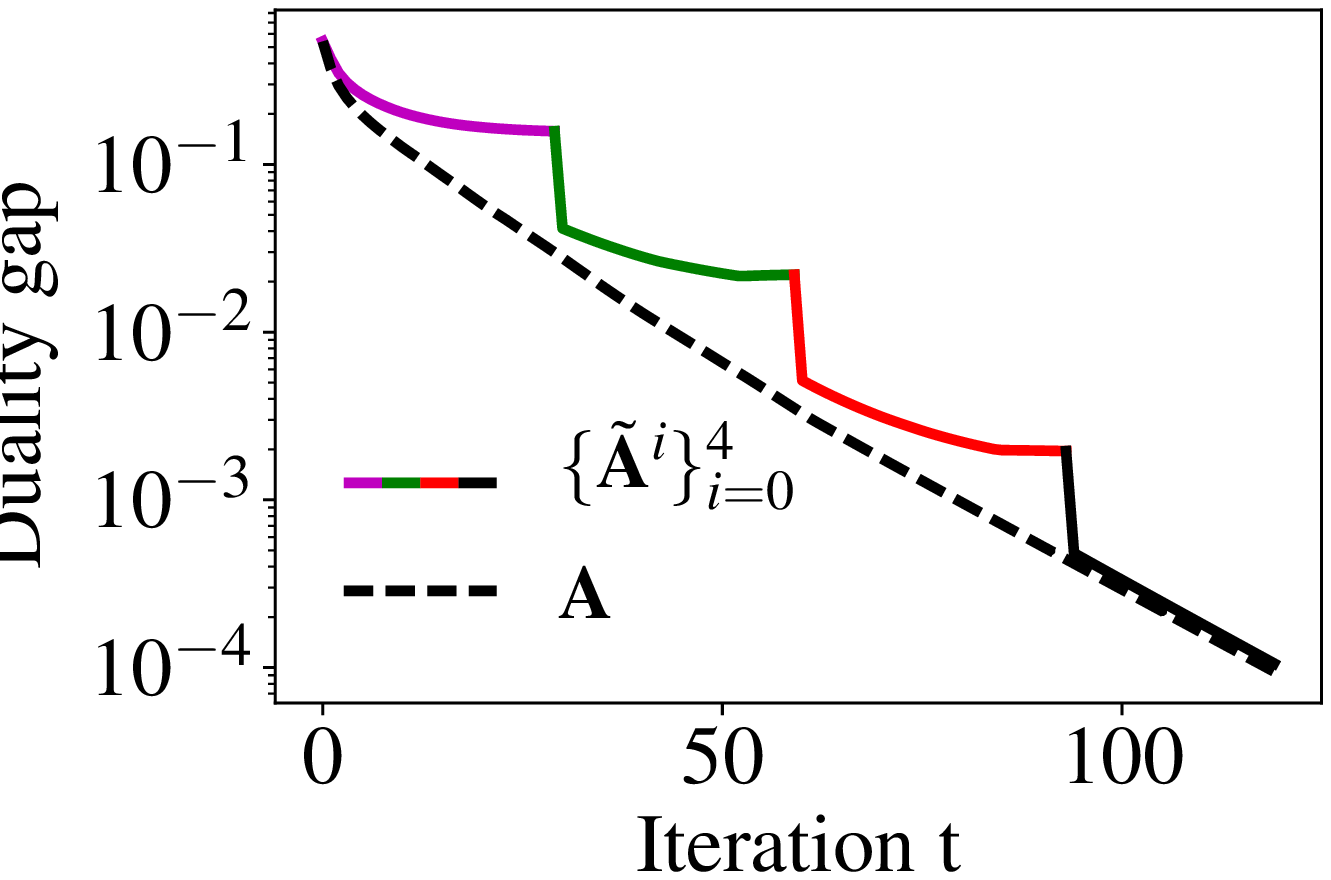}
        \caption{\label{fig:multiple_dict_gap}}
	\end{subfigure}%
	~
	\begin{subfigure}[t]{0.5\linewidth}
        \centering
        \includegraphics[width=\linewidth,trim={0 0.5cm 0 0.3cm}]{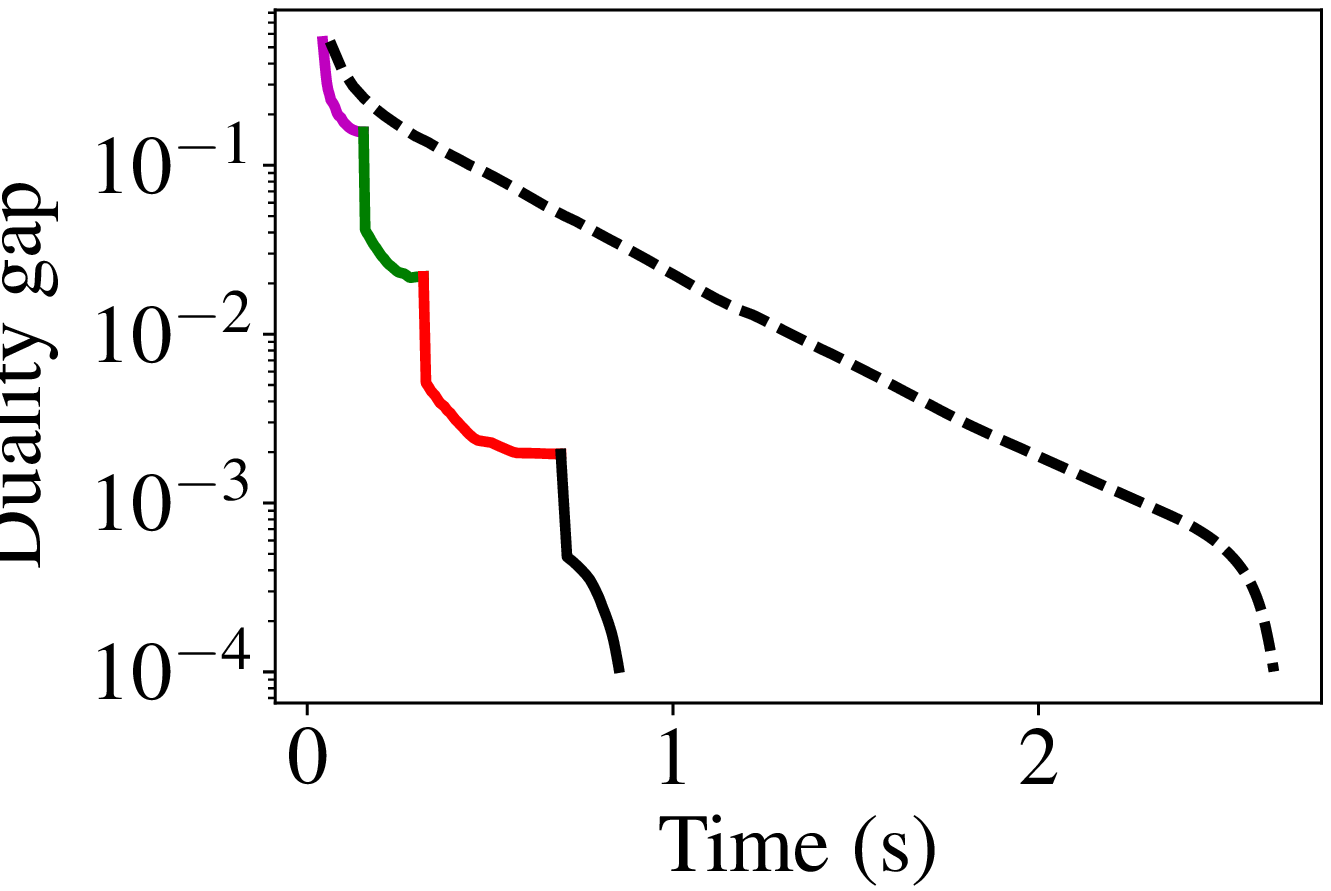}
        \caption{\label{fig:multiple_dict_gap_time}}
	\end{subfigure}
	\\
	\begin{subfigure}[t]{0.5\linewidth}
        \centering
        \includegraphics[width=\linewidth,trim={0 0.5cm 0 0.1cm}]{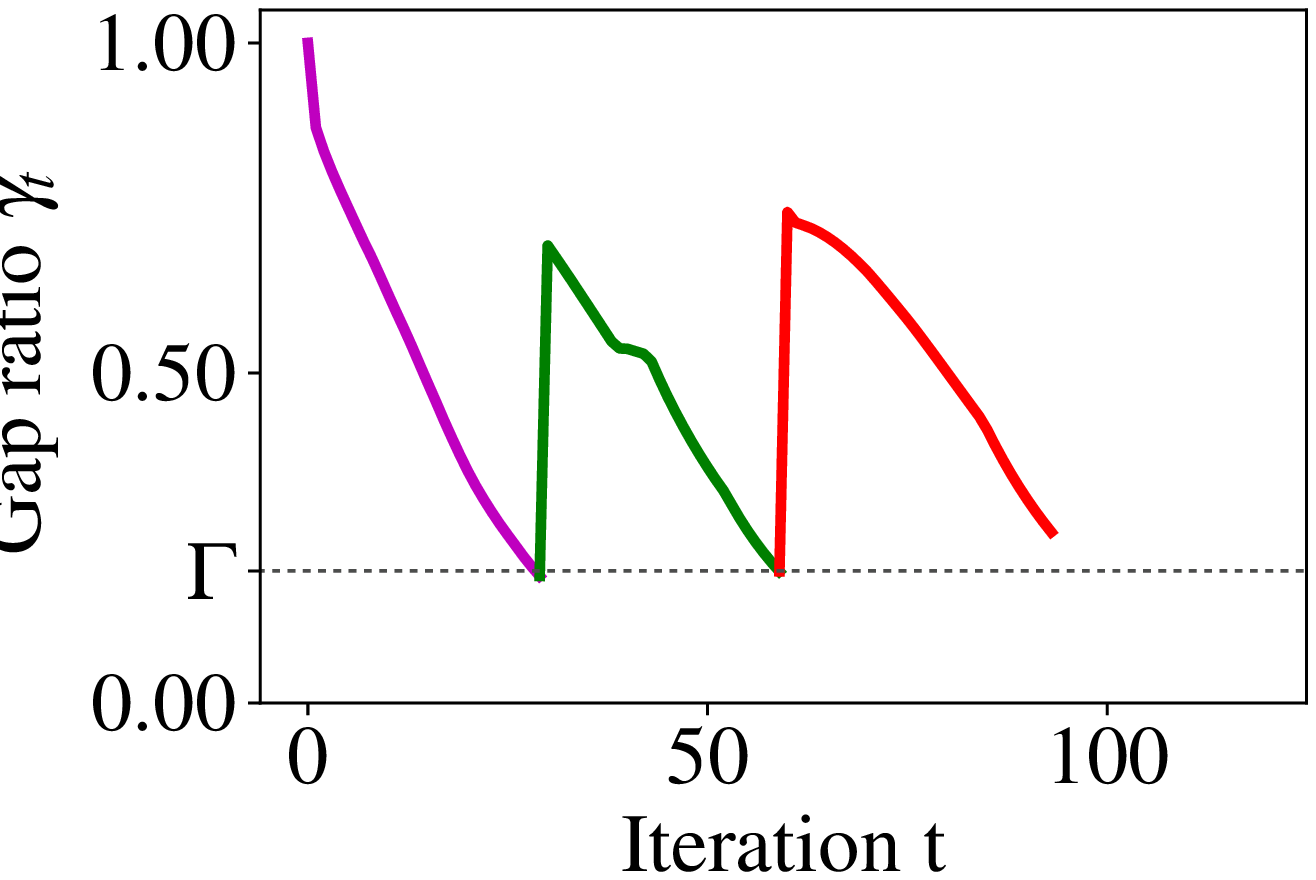}
        \caption{\label{fig:multiple_dict_ratio}}
	\end{subfigure}%
	~
	\begin{subfigure}[t]{0.5\linewidth}
        \centering
        \includegraphics[width=\linewidth,trim={0 0.5cm 0 0.1cm}]{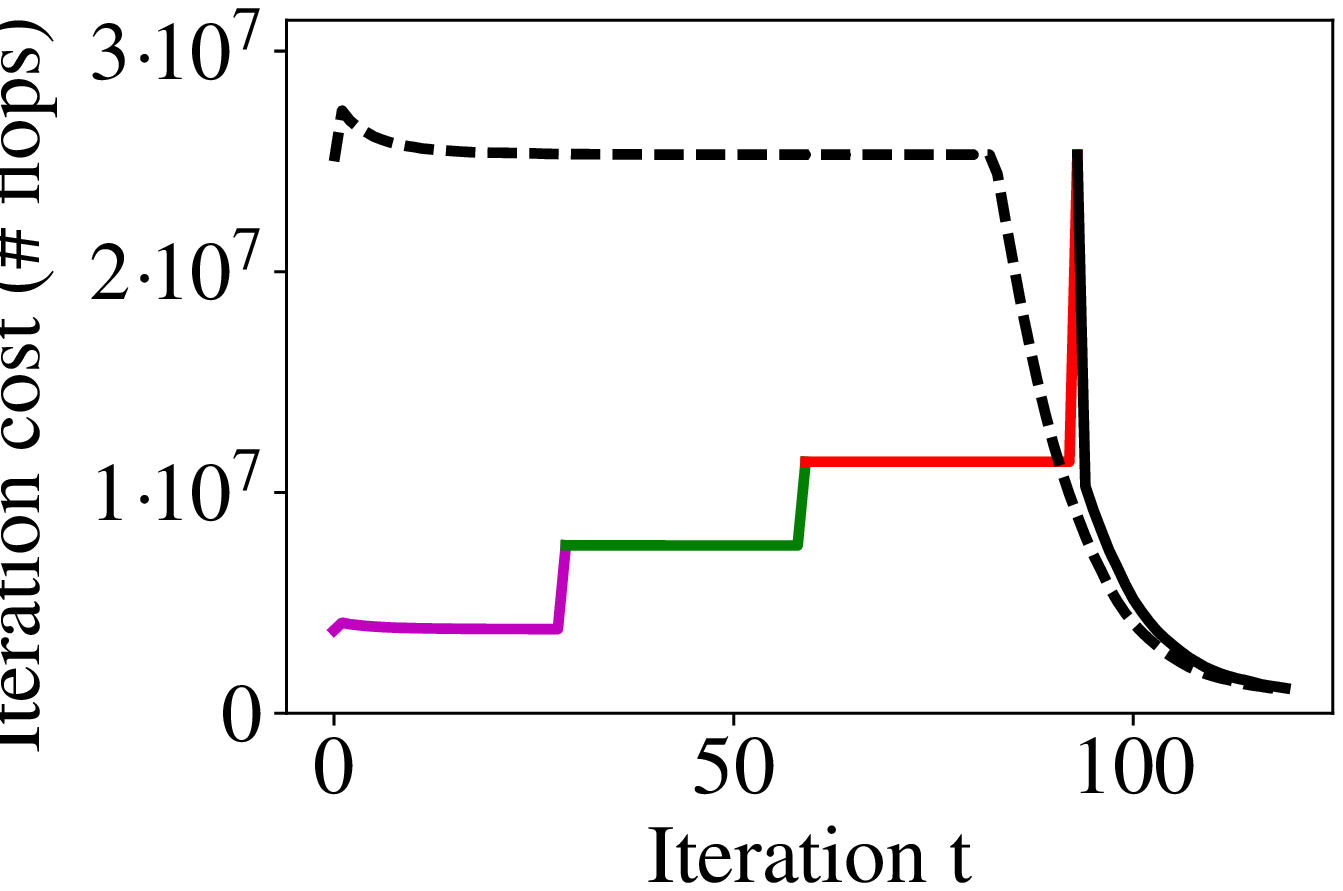}
        \caption{\label{fig:multiple_dict_flops}}
	\end{subfigure}	
\caption{Application example of the proposed algorithm. (a) and (b) Duality gap over the iterations and time respectively. (c) Gap ratio $\gamma_t$ (cf. \eqref{eq:DefRatioSwitch}) for the convergence switching criterion, with threshold $\Gamma\!=\!0.2$. (d) Computational complexity per iteration (worst-case theoretical number of flops, cf. section \ref{ssec:switching_complexity}). $N=2500$, $K=10000$, $\lambda/\lambda_{max}=0.2$.
\label{fig:multiple_dict}}
\end{figure}

\paragraph*{Example}
Figure \ref{fig:multiple_dict} shows an example of the proposed algorithm, given a list of approximations $\{\tilde{\mt{A}}^i\}_{i=0}^{I}$ with ${I=4}$,
decreasing approximation error $\vt{\epsilon}^i$ 
and increasing relative complexity $\text{RC}(\tilde{\mt{A}}^i) \!=\! 0.15({i\!+\!1})$, $0 \!\leq\! i \!<\! I$.
As usual, $\tilde{\mt{A}}^I \!=\! \mt{A}$.

Figure \ref{fig:multiple_dict_gap} shows the duality gap evolution over the iterations with $\Gamma = 0.2$. As soon as the gap $G(\vt{x}_t, \vt{\theta}'_t|\tilde{\mt{A}^i})$ saturates, a more precise approximation is adopted (the next on the list), avoiding to excessively delay the convergence compared to the conventional case where screening is performed with $\mt{A}$ from the beginning (dotted curve).
Although some delay is introduced in terms of iterations until convergence, 
it is largely compensated by the fact that the initial iterations with the approximate dictionaries are much faster. This is illustrated in Fig. \ref{fig:multiple_dict_gap_time} in which the duality gap is plotted as a function of the execution time. 

The theoretical complexity per iteration is shown in Fig. \ref{fig:multiple_dict_flops}. As better approximations are adopted, the iteration complexity rises proportionally to the corresponding RC. Note that as soon as the conventional dictionary becomes faster than the current approximation, it is promptly adopted (by the speed criterion). In this example, the last approximation $\tilde{\mt{A}}^{i=3}$ was never used, since we switched directly from $\tilde{\mt{A}}^{i=2}$ to $\mt{A}$.

\subsection{Complexity Analysis}

Existing screening tests introduce only a minor computational overhead because they primarily reuse matrix-vector multiplications either already performed in the optimization algorithm update $p(\mt{\vt{x}_t,\mt{A},\vt{\alpha}_t})$ (typically the product $\mt{A}^T\vt{\rho}$) or that can be precalculated once for all ($\mt{A}^T\vt{y}$). 
The same holds for the stable tests proposed in this article. We now derive the involved number of floating point operations (flops).
Since static screening rules represent only a fixed computational overhead, we concentrate on dynamic rules which could potentially (if not properly designed) lead to a significant overhead.

\subsubsection{Screening cost}

The most expensive computations associated to a dynamic screening rule -- the ones potentially of the order of a matrix vector product $\mathcal{O}(NK)$ -- are:
\begin{itemize}
\item Computation of a dual feasible point $\vt{\theta}_t$ in \eqref{eq:theta_dynamic} (resp. $\vt{\theta}'_{t}$\eqref{eq:theta_dynamic_approx}) requires the product $\mt{A}^T\vt{\rho}$ (resp. $\tilde{\mt{A}}^T\tilde{\vt{\rho}}$) reused from the optimization algorithm.
\item Sphere test $\test{\vt{a}}{\mathcal{R}}$ (resp. $\test{\mathcal{S}'_j}{\mathcal{R}}$) requires the product $\vt{a}^T_j \vt{c}$ (resp. $\tilde{\vt{a}}^T_j \vt{c}$) for all preserved atoms $j \in \mathcal{A}$, which comes down to the matrix-vector product $\vt{A}^T \vt{c}$ (resp. $\tilde{\vt{A}}^T \vt{c}$). 
Practical sphere regions have been designed to limit this potential overhead.
While in the Dynamic Safe sphere it reduces to the precalculated product $\mt{A}^T\vt{y}$, in the GAP Safe sphere it reduces to $\mt{A}^T\vt{\rho}$ (resp. $\tilde{\mt{A}}^T\tilde{\vt{\rho}}$) calculated in the optimization iteration.
\end{itemize}

The other required calculations are detailed in Appendix \ref{apx:complexity_screening}. 
In short, the screening represents a rather low overhead $\mathcal{O}(N+|\mathcal{A}|)$ --even its stable version-- compared to the optimization update:  $\mathcal{O}(N|\mathcal{A}|)$ with screening or $\mathcal{O}(NK)$ without it, due to matrix-vector products.

\subsubsection{Full iteration cost}

Table \ref{table:complexity_iteration} shows the number of operations of a complete iteration in Algorithm \ref{alg:ISTA_dynamic_approx} (ISTA update + screening), following~\cite{Bonnefoy2015} and
adopting the complexity models in equations \eqref{eq:complexity_model} and \eqref{eq:complexity_model_approx} for matrix vector multiplications.
We denote $\text{flops}_{\mt{A}}(t)$ the cost of iteration $t$ with the conventional screening and $\text{flops}_{\tilde{\mt{A}}}(t)$ with the stable screening. 
As a benchmark, we use the complexity of an ISTA iteration without screening, denoted $\text{flops}_N(t)$.

\begin{table}[h]
\renewcommand{\arraystretch}{1.3} 
\centering
	\begin{tabular}{|c|c|}
	\hline
	$\text{flops}_N(t)$ & $(K + \|\vt{x}_t\|_0)N + 4K + N$ \\ \hline
	$\text{flops}_{\mt{A}}(t)$ & $(|\mathcal{A}| + \|\vt{x}_t\|_0)N + 6|\mathcal{A}| + 5N$ \\ \hline
	$\text{flops}_{\tilde{\mt{A}}}(t)$ & $(\text{RC} \!\times\! K + \|\vt{x}_t\|_0)N + 8|\mathcal{A}| + 7N$	 \\
	\hline
	\end{tabular}
\caption{Complete iteration complexity}\label{table:complexity_iteration}
\end{table}

To obtain the total complexity of the algorithm, 
simply add up all iteration costs calculated with the corresponding active set size $|\mathcal{A}_t|$ and sparsity of the solution estimate $\|\vt{x}_t\|_0$.

\section{Experiments} \label{sec:experiments}

In this section we demonstrate the potential of the proposed algorithm in terms of complexity reduction and time saving for $\ell_1$-minimization problems. This is done in a wide set of simulation scenarios, summarized  in Table \ref{table:simulation_parameters}, to evaluate the influence of the main  parameters involved.

\begin{table}[h]
\renewcommand{\arraystretch}{1.3} 
\centering
	\begin{tabular}{|c|c|}
	\hline
	\bf{Problem parameters} & \bf{Values} \\ \hline\hline
	Regularization $\lambda/\lambda_{max}$ & $[10^{-2},1]$ (logarithmically-spaced)\\
	Convergence tolerance & $G(\vt{x}_t,\vt{\theta}_t|\mt{A})< tol \in [10^{-6},10^{-3}]$ \\
	Data distribution $\vt{y}$ & $\mt{A} \vt{x}$ with $\vt{x}$ Bernoulli-Gaussian \\ 
	\hline \hline
	\bf{Algorithm parameters} & \bf{Values} \\ \hline\hline
	Optimization algorithm & ISTA, FISTA \\
	Screening Test & (stable) Dynamic, (stable) GAP Safe \\
	Switching threshold $\Gamma$ & $[0.01, 0.8]$ (logarithmically-spaced)\\
	Speedup-Error tradeoff $\tilde{\mt{A}}^i$ & 3 scenarios (see Section \ref{ssec:SpecifyingStructured}) \\ 
	\hline
	\end{tabular}
\caption{Pool of parameters explored in the simulations.}\label{table:simulation_parameters}	
\end{table}

The dimension of the observed vector $\vt{y}$ is $N=2500$ and that of the  coefficient vector $\vt{x}$ (or, equivalently, the number of atoms in the dictionary) is $K=10000$.
Unit-norm input data samples $\vt{y}=\mt{A}\vt{x}$ are generated from a sparse vector $\vt{x}$ with active set determined by a Bernoulli distribution with probability $p=0.02$ and zero-mean standard Gaussian entries.
The reported results are the average on 25 independent and identically distributed realizations of the input $\vt{y}$.
The dictionary matrix $\mt{A} \in \mathbb{R}^{N \times K}$ is generated in such a way that it is more or less efficiently approximated by a fast structured matrix according to three representative scenarios (more details are given in Section \ref{ssec:SpecifyingStructured}).

Although other problem dimensions ($N$ and $K$) and data distributions were explored, we decided to keep these parameters 
fixed on the reported experiments since they were observed not to decisively influence the analyzed results. 

In all figures, the GAP Safe and Dynamic Safe tests are denoted respectively by the acronyms \textbf{GAP} and \textbf{DST}.

\subsection{Specifying the Fast Structured dictionaries} \label{ssec:SpecifyingStructured}

The performance of the proposed algorithm is directly associated to the \emph{quality} of the available approximations $\tilde{\mt{A}}^i$. A good approximation would have both a small application complexity (RC) and small approximation error ($\vt{\epsilon}$). There is a compromise, however, since each of these features usually come to the price of the other.

In the experiments, we use a particular kind of fast structured dictionaries referred to as \emph{SuKro}~\cite{Dantas2017} which can be written as a sum of Kronecker products $\tilde{\mt{A}} \!=\! \sum_{k=1}^{n_{\text{kron}}} \mt{B}_k \!\otimes\! \mt{C}_k$.
Its reduced multiplication cost comes from the fact that the sub-matrices $\mt{B}_k$, $\mt{C}_k \in \mathbb{R}^{\sqrt{N} \!\times\! \sqrt{K}}$ are much smaller than $\mt{A} \in \mathbb{R}^{N \!\times\! K}$.
The choice of the SuKro structure is justified by the fact that it directly provides a range of speedup-error compromises by tweaking the number of Kronecker terms ($n_{\text{kron}}$) in the sum. A higher $n_{\text{kron}}$ provides a more precise approximation although implying a higher RC.

\begin{figure}
    \centering
    \includegraphics[width=0.6\linewidth,trim={0 0.4cm 0 0.4cm}]{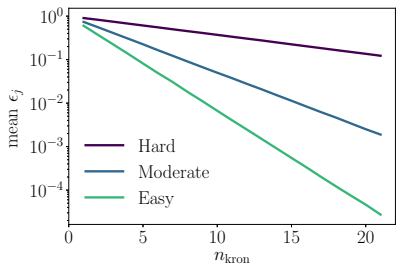}
    \caption{Simulated scenarios: different trade-offs between approximation error and speedup of the structured dictionaries. 
\label{fig:simulation_scenarios}}
\end{figure}

We define three representative simulation scenarios -- easy, moderate and hard scenarios -- in which the dictionary matrix is poorly, moderately and efficiently approximated by the structured dictionary respectively.
In practical terms, this translates to different approximation error decay profiles as a function of the number of Kronecker terms ($n_{\text{kron}}$) on the SuKro structure, as shown in Figure \ref{fig:simulation_scenarios}.
The easier the scenario, the faster the approximation error decays as a function of $n_{\text{kron}}$.

Although computational complexity associated to a certain SuKro operator can be calculated analytically, 
we measured the actual time speedup obtained in practice in order to have a more realistic value.
Interestingly, the measured RC is lower than the theoretical prediction.
\footnote{Some parallelization is introduced (in the summing terms) and matrix-matrix products are faster than the naive cubic complexity.}

\begin{table}[h]
\renewcommand{\arraystretch}{1.3} 
\centering
	\begin{tabular}{|c|c|c|c|c|}
	\hline
	$n_{\text{kron}}$		&\bf{5}	&\bf{10}&\bf{15}&\bf{20}\\ \hline \hline
	\textbf{Measured RC}	& 0.08 	& 0.15 	& 0.21 	& 0.28 	\\ \hline
	\textbf{Theoretical RC}	& 0.15 	& 0.30 	& 0.45 	& 0.60 	\\ \hline
	\end{tabular}
\caption{Relative Complexities (RC)}\label{table:complexity_iteration}
\end{table}

\subsection{Computational complexity and Time results}
Let us denote $F_N$, $F_{\mt{A}}$ and $F_{\tilde{\mt{A}}}$ respectively the total computational complexity (in number of flops) of the optimization algorithm without screening, with the conventional screening  and with the proposed approach, such that
\begin{align}
F_X = \sum_{t=1}^{n_\text{it}} \text{flops}_X (t), \quad X \in \{N,\mt{A},\tilde{\mt{A}}\}
\end{align}
where $n_\text{it}$ is the number of iterations. 
When calculating $F_{\tilde{\mt{A}}}$ , the current approximation at iteration $t$ must be used in the expression of $\text{flops}_{\tilde{\mt{A}}}(t)$.
Likewise, we denote $T_N$, $T_{\mt{A}}$ and $T_{\tilde{\mt{A}}}$ the measured running times.

As a main figure of merit, we evaluate the normalized number of flops 
($F_{\mt{A}}/F_{N}$ and $F_{\tilde{\mt{A}}}/F_{N}$) 
and normalized running times ($T_{\mt{A}}/T_{N}$ and $T_{\tilde{\mt{A}}}/T_{N}$)
\footnote{Time results are obtained in an 
Intel\textsuperscript{\textregistered} Core\texttrademark{} i7-5600U CPU @ 2.60GHz, 16GB RAM.
But since mostly time ratios are reported, the results here should be relatively consistent to other machine specifications.
}.

\begin{figure}
	\centering
    \includegraphics[width=0.79\linewidth,trim={0 0.5cm 0 0.3cm}]{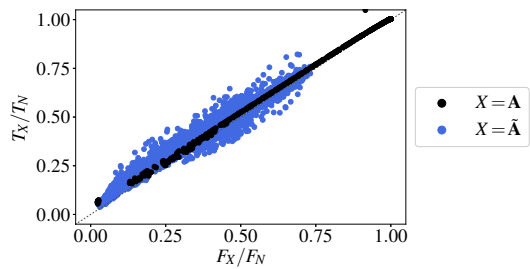}
        \caption{Normalized number of flops vs. normalized running times for ISTA ($tol\!=\!10^{-5}$, $\Gamma\!=\!0.5$). Each point on the graph corresponds to an independent run at a given regularization $\lambda/\lambda_{\max} \!\in\! [10^{-2},1]$. Results for the usual GAP Safe test are shown in black and those using its stable version (as in Alg. \ref{alg:ISTA_dynamic_approx}) are shown in blue. The closer the dots lie to the identity line, the better the theoretical speedups are met in practice. \label{fig:complexity_time_comparison}}
\end{figure}

In all simulated scenarios, the observed time reductions match closely the theoretical speedups predicted in terms of computational complexity, as illustrated in Figure \ref{fig:complexity_time_comparison} for the GAP Safe rule.
It shows the correlation between theoretical complexities and measured running times in multiple independent runs. 
The fact that the points are well concentrated around the identity line is an empirical evidence that the predicted speedups really translate into practical accelerations.
Similar results are obtained for the Dynamic Safe rule, for other convergence tolerances and switching thresholds.
Given this observation, in the remainder of the paper we report only running time results, which are more relevant in practice.

\subsection{Choosing the switching threshold} \label{ssec:Choosing_gamma}

The convergence-based switching criterion proposed in Section \ref{ssec:switching_convergence} relies on a hyperparameter: the threshold $\Gamma$. As discussed then, this parameter determines how long the approximate dictionaries are kept. 

\begin{figure}
	\centering
    \includegraphics[width=\linewidth,trim={0 0.5cm 0 0.3cm}]{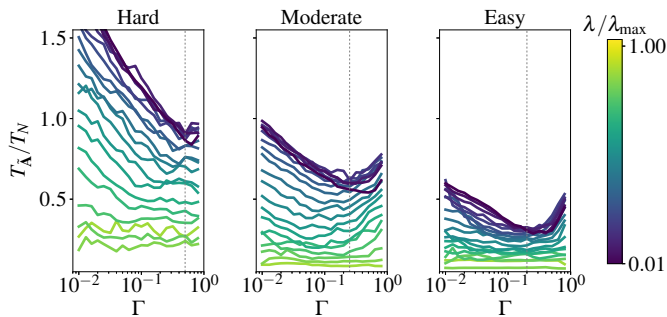}
        \caption{Impact of the switching threshold $\Gamma$ (x-axis) on the time results (y-axis) of Algorithm \ref{alg:ISTA_dynamic_approx} using FISTA and GAP Safe. Each line corresponds to a regularization $\lambda/\lambda_{\max} \!\in\! [10^{-2},1]$. \label{fig:choosing_gamma}}
\end{figure}

We empirically observed that the choice of $\Gamma$ is mostly dependent on the quality of the approximations $\tilde{\mt{A}}^i$ (represented here by the three simulation scenarios defined in Section \ref{ssec:SpecifyingStructured}). 
In Figure \ref{fig:choosing_gamma} we show the normalized times as a function of $\Gamma \in [10^{-2}, 0.8]$ for each one of the three scenarios. Each line corresponds to a different regularization level $\lambda$.
For each of these lines, we are interested in the $\Gamma$ value that minimizes the running times.
We can see that the low regularization configurations (dark blue curves) are more sensitive to choice of $\Gamma$, especially in the Hard scenario
(left plot) in which a bad choice of $\Gamma$ can even lead to normalized running times greater than one --so, actually, a slowdown. This happens when a too small $\Gamma$ is used, which means that the approximate dictionaries are kept longer than they should, causing an important detour in the convergence path and thus delaying convergence. 

There is no reason for the optimal $\Gamma$ to be the same for every regularization level.
However,
a common behavior is observed regardless of the regularization and, for a given simulation scenario, 
a single $\Gamma$ value can be chosen to obtain close to optimal execution times for any $\lambda/\lambda_{\max}$ in the tested range.  
Slightly different $\Gamma$ values can be chosen to better adapt to each simulation scenario:
$\Gamma = 0.5, 0.25, 0.2$ (indicated by the dotted vertical lines in the figure) respectively on the hard, moderate and easy scenarios.
In general terms, the \emph{worse} the available approximations are (i.e. the harder the scenario) the higher the $\Gamma$ to be chosen. This is intuitive, since it implies being more conservative in switching earlier when the available approximations are of lower quality.

If the aimed speedup-error compromise is not fixed, a good general compromise is picking $\Gamma = 0.5$, for instance. 

\subsection{Single vs. Multiple Approximations}

In our experiments, using multiple dictionary approximations proved to be always advantageous when compared to using a single approximation, as soon as the switching parameter $\Gamma$ is well-tuned (see Section \ref{ssec:Choosing_gamma}).

\begin{figure}[t]
    \centering
    \includegraphics[width=\linewidth,trim={0 0.5cm 0 0.4cm}]{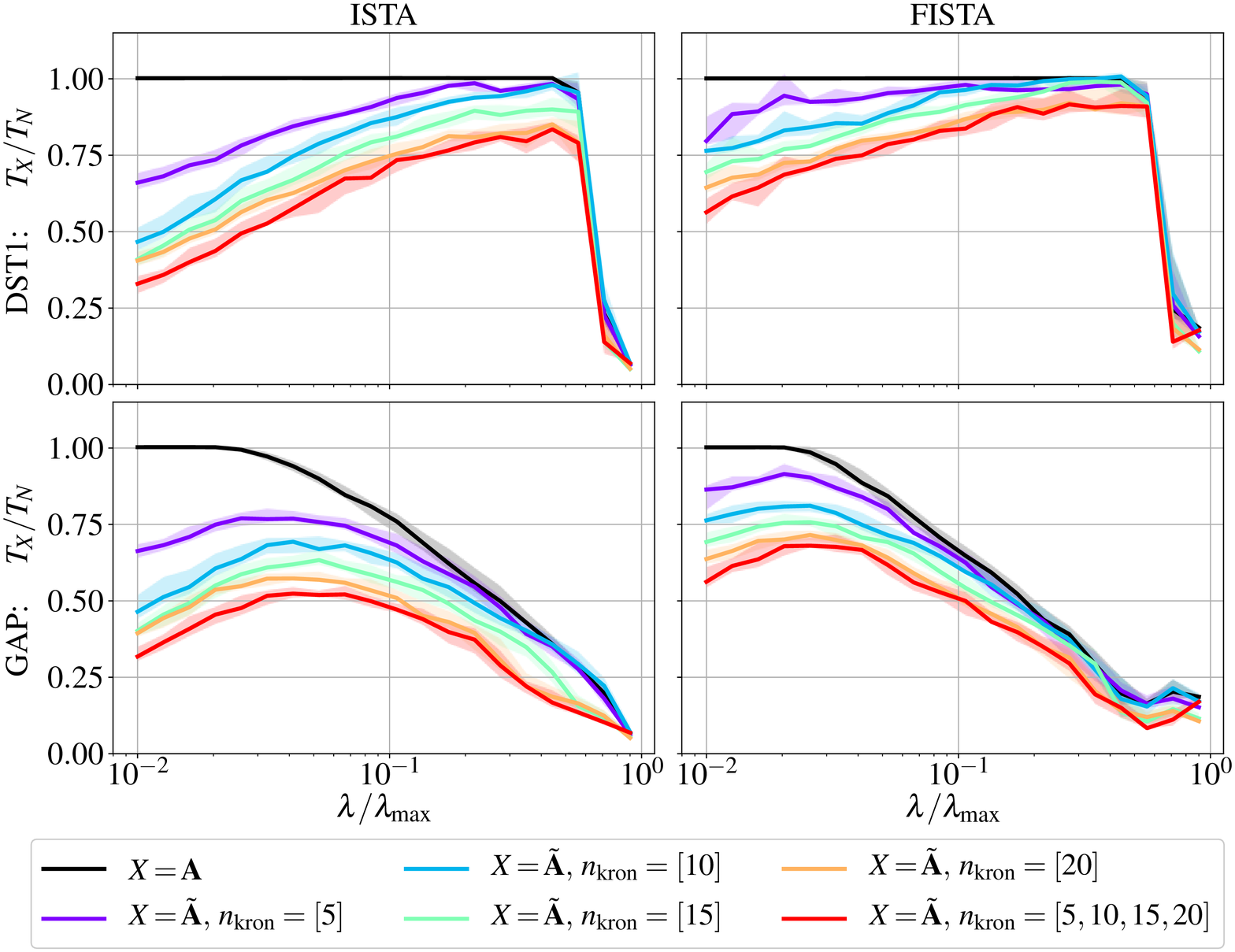}
    \caption{Running times normalized w.r.t. $T_{N}$ (ISTA or FISTA algorithms without screening). Conventional and Stable screening results respectively in black and colored lines. Left: ISTA. Right: FISTA. Top: Dynamic Safe screening. Bottom: GAP Safe. Moderate scenario, $tol\!=\!10^{-5}$,  $\Gamma\!=\!0.5$.\label{fig:single_vs_multiple}}
\end{figure}

Figure \ref{fig:single_vs_multiple} shows the normalized running times for an entire range of regularizations and corroborates the previous statement for both GAP and Dynamic Safe tests as well as for both ISTA and FISTA algorithms. The medians among 25 runs are plotted and the shaded area contains the 25\%-to-75\% percentiles.
Although the single-approximation version of the proposed algorithm (as introduced in \cite{F.Dantas2017,F.Dantas2018}) already provides noticeable speedups to the tested gradient-based solvers, even 
with respect to their screening-based implementation (black curves), the use of multiple approximations (red curves) consistently leads to even better speedups.
This ratifies the relevance of the generalization to multiple dictionaries introduced in this paper, by relying on more robust switching criteria.
Similar results, which are omitted here, were obtained for other convergence tolerances and speedup-error compromise (these parameters are further explored in Sections \ref{ssec:Exp.Speedup-error} and \ref{ssec:Exp.Conv-tolerance}).

A smaller speedup is provided to the FISTA algorithm when compared to ISTA, which is expected since the former is already faster than the latter\footnote{The results reported in the right graphs are already normalized w.r.t the FISTA time results, which are typically smaller than those of ISTA.}.
A similar argument applies to the GAP Safe rule, which has stronger screening capabilities than Dynamic Safe:
our method still manages to provide some additional acceleration when combined to these already quite efficient techniques, especially in low-regularized scenarios.

For simplicity, only the results concerning multiple dictionary approximations will be reported from this point on.

\subsection{Speedup-error compromise} \label{ssec:Exp.Speedup-error}

\begin{figure}[t]
    \centering
    \includegraphics[width=\linewidth,trim={0 0.5cm 0 0.4cm}]{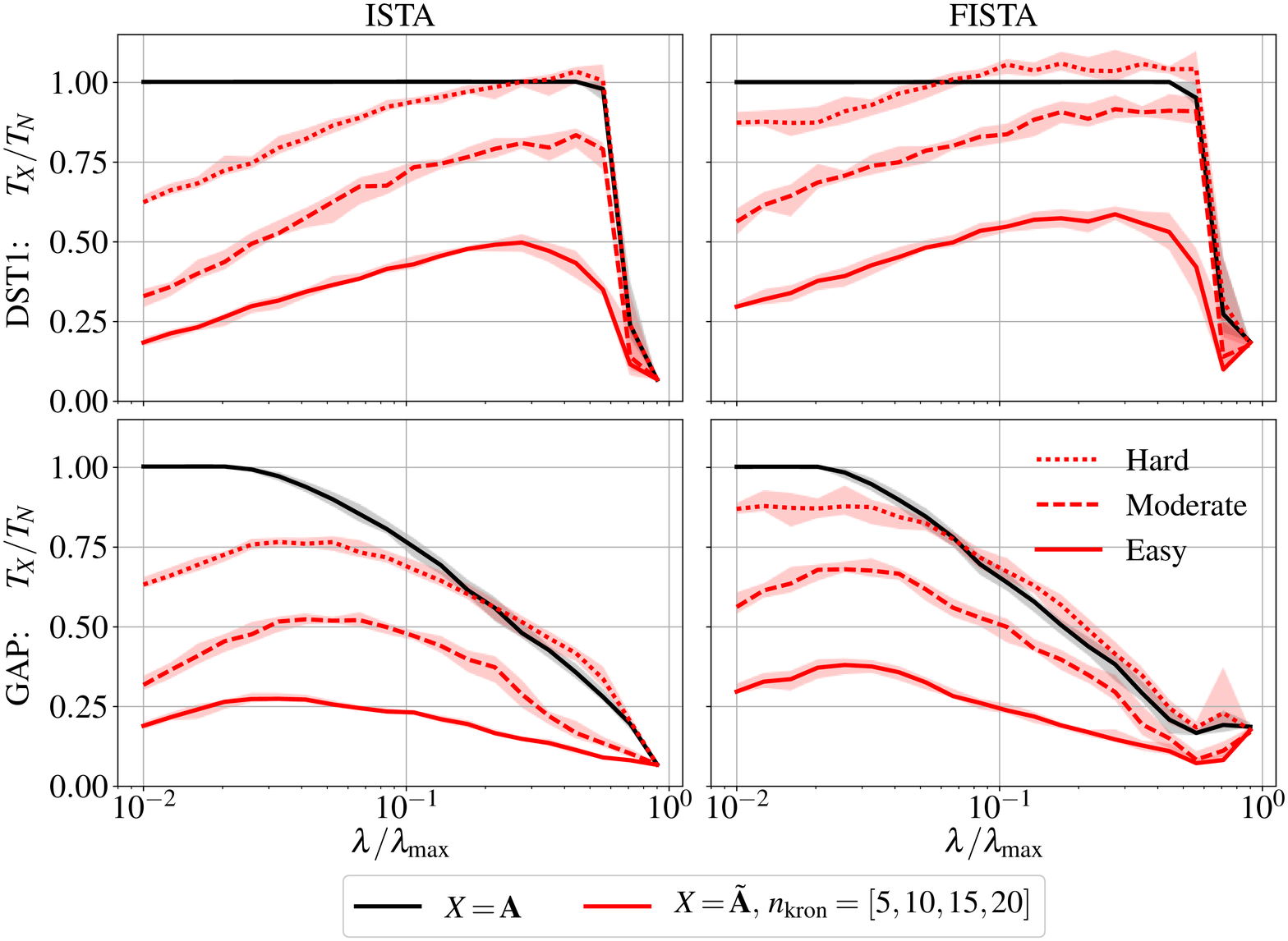}
    \caption{Normalized running times in 3 different scenarios of speedup-error tradeoffs. $tol\!=\!10^{-5}$,  $\Gamma\!=\!0.5$.\label{fig:speedup-error_compromise}}
\end{figure}

It is reasonable to expect the quality of the approximations $\tilde{\mt{A}}^i$ to be decisive on the performance of the proposed algorithm. Its success depends on the possibility of providing fast yet precise approximations of the dictionary $\mt{A}$. Let us now evaluate the influence of this speedup-error compromise on the simulation time results.

Figure \ref{fig:speedup-error_compromise} shows the same type of normalized running times as in Figure \ref{fig:single_vs_multiple}, but it now compares the results of the proposed algorithm under three different scenarios in terms of speedup-error profiles (see 
Figure \ref{fig:simulation_scenarios} for details on the approximation errors in each scenario). The impact on the algorithm performance is indeed significant. However, even on the Hard scenario, the proposed approach manages to provides non-negligible acceleration.

\subsection{Convergence tolerance} \label{ssec:Exp.Conv-tolerance}

\begin{figure}
	\centering
	\begin{subfigure}[t]{1.0\linewidth}
        \centering
        \includegraphics[width=\linewidth,trim={0 0.5cm 0 0.3cm}]{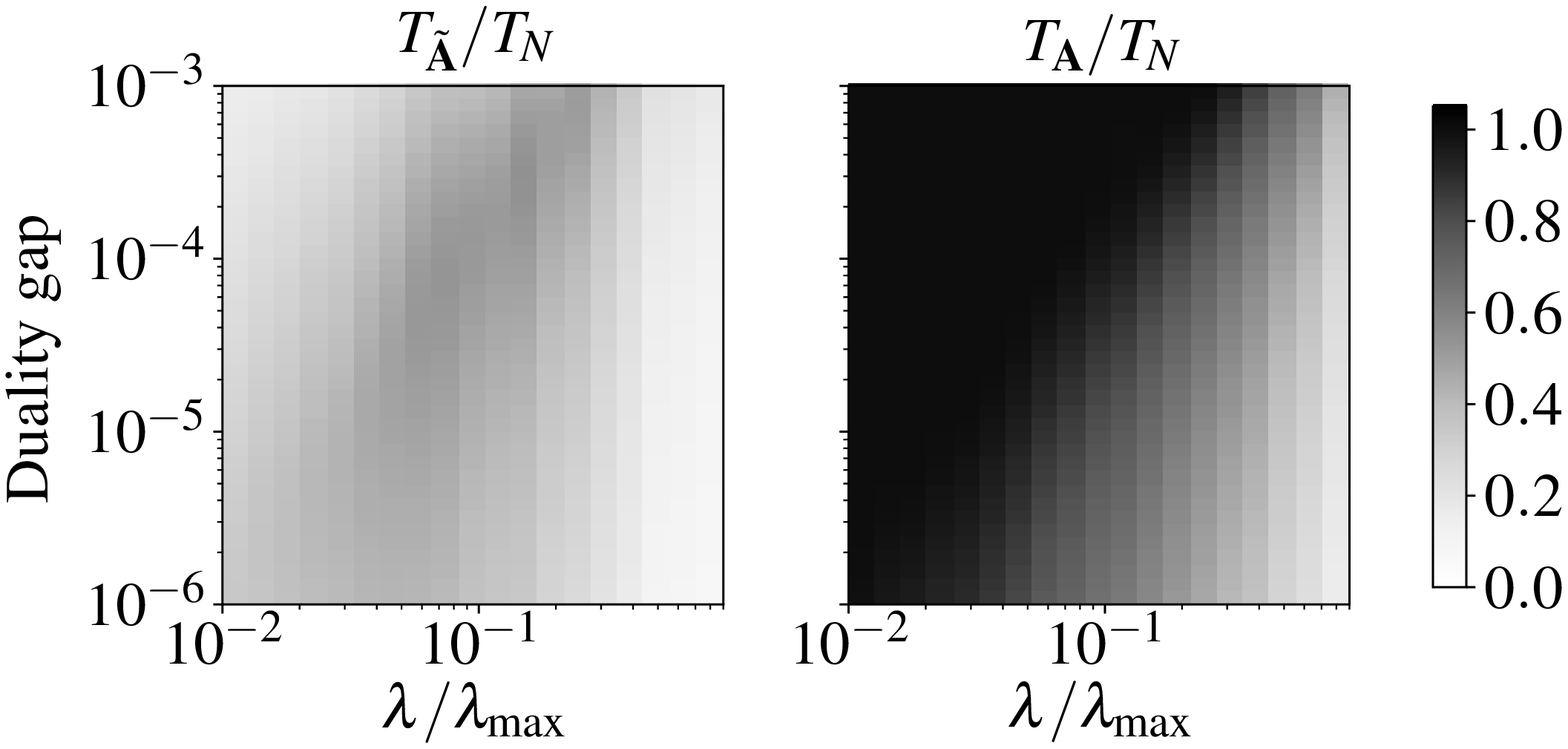}
        \caption{ISTA\label{fig:tolerance_colormap_ISTA}}
	\end{subfigure}
	\\
	\begin{subfigure}[t]{1.0\linewidth}
        \centering
        \includegraphics[width=\linewidth,trim={0 0.5cm 0 0.1cm}]{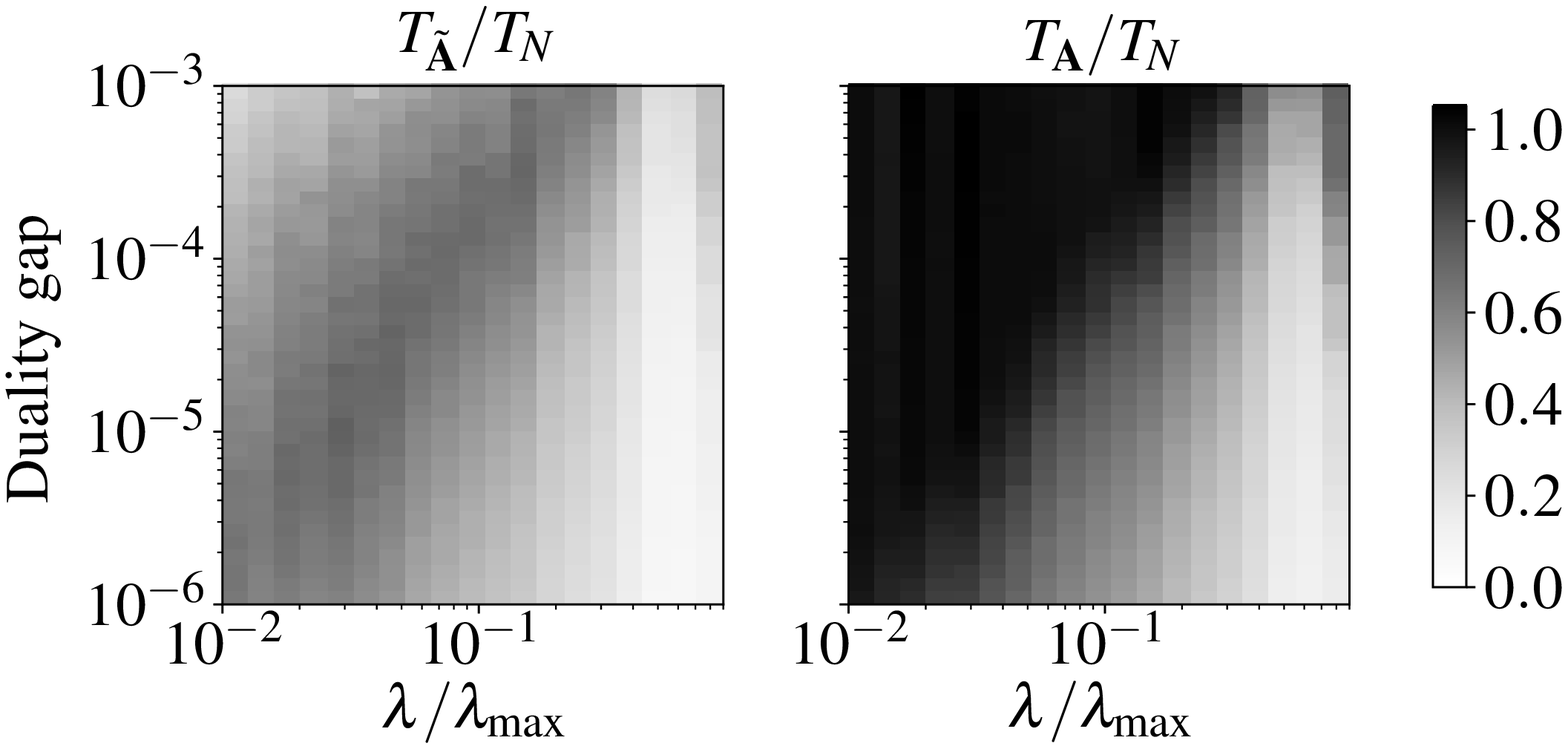}
        \caption{FISTA\label{fig:tolerance_colormap_FISTA}}
	\end{subfigure}
    \caption{Normalized execution times in grayscale (darker means slower) to reach a certain duality gap  for a range of regularizations. Left: proposed technique (w/ stable screening). Right: Conventional screening.\label{fig:tolerance_colormap}}
\end{figure}

As previously highlighted, screening is known to work better on highly-regularized scenarios ($\lambda/\lambda_{\max} \approx 1$). The speedup provided by the screening is also more pronounced when a more precise convergence is required, because the final iterations are often less costly as most atoms have been screened out. These two aspects create the triangular profiles observed in Figure \ref{fig:tolerance_colormap} (right plots). 
This figure shows the normalized execution times in grayscale (the darker, the slower) as a function of both the regularization (x-axis) and the convergence precision in terms of the duality gap (y-axis).

Note that the proposed method (left plots) efficiently complements the screening tests in its main weaknesses, namely: low regularizations and reduced convergence requirements. 
In weakly regularized scenarios, while screening tests struggle to eliminate atoms especially in initial iterations, the fast approximate dictionaries come at rescue by making those iterations faster.
Besides that, when higher duality gaps are targeted, less iterations using the slow dictionary $\mt{A}$ are necessary. Thus, proportionally, the accelerated part of the algorithm (with fast dictionaries) is more significant.

\subsection{MEG source localization}\label{ssec:real}

In this experiment, we consider an MEG (magnetoencephalography)  source localization problem,
which consists in retrieving a limited set of active brain foci associated to a given MEG measured signal. A common way to achieve such a task is to solve a convex sparse inverse problem \cite{Matsuura1995,Gramfort2012}. 

We consider an MEG gain matrix (forward operator) $\mt{A} \!\in\! \mathbb{R}^{204\!\times\!8193}$ obtained with the MNE software \cite{Gramfort2014}. 
This operator can be efficiently approximated as a product of a few sparse matrices, as proposed in \cite{Magoarou2016}. 
The speedups and corresponding approximation errors achieved by the so-called \emph{FA$\mu$ST} dictionaries adopted here are summarized in Table \ref{table:MEG_approx-error_RC}.
The obtained speedup-error compromise corresponds to a hard scenario. 

The experiment consists in picking (uniformly) at random eight active sources with standard gaussian weights giving \mbox{8-sparse} coefficient vectors $\vt{x} \in \mathbb{R}^{8193}$ to be recovered from the input signals $\vt{y}\!=\!\mt{A}\vt{x} \in \mathbb{R}^{204}$ by solving problem \eqref{eq:lasso}.

\begin{table}[h]
\renewcommand{\arraystretch}{1.3} 
\centering
	\begin{tabular}{|c|c|c|c|c|}
	\hline
	\textbf{Mean $\vt{\epsilon}_j$}	& $2.1\!\cdot\!10^{-1}$ &  $1.2\!\cdot\!10^{-1}$ &  $6.6\!\cdot\!10^{-2}$ & $5.9\!\cdot\!10^{-2}$ \\ \hline
	\textbf{Measured RC}	& 0.22 	& 0.33 	& 0.45 	& 0.63 	\\ \hline
	\end{tabular}
\caption{Approximation errors and Relative Complexities for the FAuST approximations of the MEG matrix.}\label{table:MEG_approx-error_RC}
\end{table}

\begin{figure}
	\centering
    \includegraphics[width=\linewidth,trim={0 0.5cm 0 0.0cm}]{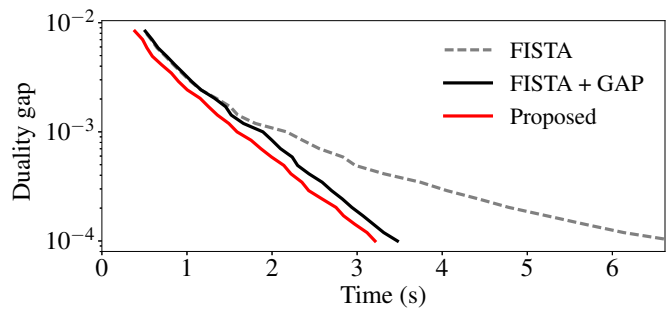}
	\caption{Source localization using an MEG forward operator. Duality gap as a function of time for $\lambda/\lambda_{\max}= 10^{-1}$. \label{fig:MEG_GAP-Time}}
\end{figure}

Figure \ref{fig:MEG_GAP-Time} shows the required time to reach a certain duality gap precision (averaged over 50 independent runs) for  a fixed regularization $\lambda/\lambda_{\max} \!=\! 10^{-1}$, around which the correct number of sources is recovered.
The proposed method is consistently advantageous, providing some speedup with respect to both the FISTA algorithm (grey dotted line) and its enhanced version with GAP Safe screening (black line). 
Although somewhat modest when compared to the latter, 
the provided gain is proportionally more pronounced at lower precision requirements in the duality gap.%
\footnote{
Neither of the two considered acceleration techniques affects the algorithms results in terms of source localization accuracy, since they solve the exact same problem at the same convergence precision. 
That is why task-specific performance measures are not discussed (see \cite{Gramfort2012} for more details on the performance of $\ell_1$-minimization techniques for brain source localization). 
}
More significant speedups would require \emph{better} approximations of MEG gain matrix (in the sense of speedup-error compromise).

\section{Conclusion}

Aiming to combine safe screening tests and fast structured operators in an effort to accelerate the solution of $\ell_1$-minimization problems, we proposed a methodology for defining safe screening tests despite an inaccurate knowledge 
of the explanatory variables (dictionary atoms).
The proposed methodology was exemplified on some existing screening tests (stable and dynamic).
The resulting \emph{stable screening tests} were then employed in a fast algorithm, which exploits a series of fast approximations of the dictionary matrix.
Simulation results demonstrated the two combined strategies to be quite complementary, justifying the effort to conciliate them.

In terms of longer term perspectives, the first part of the paper also provides a better understanding on which error measures are more significant to the stable screening tests, potentially leading to more suited optimization criteria in structured dictionary approximation techniques. 
In a broader sense, the proposed stable tests can be seen as a robust screening tool for inaccurately-defined $\ell_1$-minimization problems in which the imprecision need not be due to the use of fast approximations
--for instance, due to intrinsic imprecisions on the measurement matrix of a given inverse problem.
In principle, the proposed framework can also be applied to other types of safe regions and error bounds 
than the ones explored in this paper, such as dome tests~\cite{Xiang2016}.
Finally, an interesting perspective is to extend these zone-based screening tests to off-the-grid generalizations of $\ell_1$-regularized problems~\cite{Bredies2013}.

\appendix
\subsection{Proof of Equation \eqref{eq:sphere_test}}
\label{apx:sphere}
Screening test for a $\ell_p$-ball, i.e. $\mathcal{R} =  B_p(\vt{c},R)$:
\begin{align} \label{eq:sup_lp-ball}
\sup_{\vt{\theta} \in B_p(\vt{c},R)}  |\vt{a}^T \vt{\theta}|
&= \sup_{\vt{u} \in B_p(\vt{0},1)}  |\vt{a}^T(\vt{c} + R\vt{u})| \nonumber \\ 
& = |\vt{a}^T\vt{c}| + R \sup_{\vt{u} \in B_p(\vt{0},1)} |\vt{a}^T\vt{u}| \nonumber \\
& = |\vt{a}^T\vt{c}| + R \|\vt{a}\|_{p^*}
\end{align}
using the definition of dual norm $\|\vt{a}\|_{p^*} \!:=\! \sup_{\vt{u} \in B_p(\vt{0},1)}|\vt{a}^T\vt{u}|$.

\subsection{Proof of Equation \eqref{eq:sphere_test_approx}}
\label{apx:stable_sphere}
From Definition \ref{def:screening_test_approx} and the  conventional sphere test in eq.~\eqref{eq:sphere_test}, the stable sphere test is given by
\begin{align} \label{eq:apx_sphere_test_approx}
\test{\mathcal{S}}{B_p(\vt{c},R)}
&= \sup_{\vt{a} \in \mathcal{S}}  \left( |\vt{a}^T\vt{c}| + R\|\vt{a}\|_{p^*} \right) \nonumber \\
&\leq \sup_{\vt{a} \in \mathcal{S}} |\vt{a}^T\vt{c}| +  R \sup_{\vt{a} \in \mathcal{S}}\|\vt{a}\|_{p^*}
\end{align}

Taking $\mathcal{S} = B_q(\tilde{\vt{a}},\epsilon)$, 
the first term gives (similarly to \eqref{eq:sup_lp-ball})
\begin{align} \label{eq:apx_sphere_test_approx_2}
\sup_{\vt{a} \in \mathcal{S}} |\vt{a}^T\vt{c}| 
&= \sup_{\vt{u} \in B_q(\vt{0},1)}  |(\tilde{\vt{a}} + \epsilon \vt{u})^T \vt{c} | 
= |\tilde{\vt{a}}^T\vt{c}| + \epsilon \|\vt{c}\|_{q^*}
\end{align}
and the last term gives
\begin{align} \label{eq:apx_sphere_test_approx_3}
\sup_{\vt{a} \in B_q(\tilde{\vt{a}},\epsilon)}\|\vt{a}\|_{p^*} 
&= \sup_{\vt{u} \in B_q(\vt{0},1)}\|(\tilde{\vt{a}} + \epsilon\vt{u}) \|_{p^*} \nonumber \\
&= \|\tilde{\vt{a}}\|_{p^*} + C \epsilon  
\end{align}
with a constant $C = \sup_{\vt{u}\in B_	q(\vt{0},1)}\|\vt{u}\|_{p^*} = N^{\left(\frac{1}{p^*} - \frac{1}{q}\right)_{+}}$, which results from the Holder's inequality.
Substituting \eqref{eq:apx_sphere_test_approx_3}  and \eqref{eq:apx_sphere_test_approx_2} in \eqref{eq:apx_sphere_test_approx} gives the bound in equation \eqref{eq:sphere_test_approx}.

\subsection{Proof of Lemma \ref{lemma:theta_dynamic_approx}} \label{apx:2}

By definition we have $\vt{\theta}_{F} := \Theta'(\vt{z}|\tilde{\mt{A}},\vt{\epsilon}) = \alpha \vt{z}$ where
\begin{align*} 
0 \leq \alpha \leq \frac{1}{\max_j \left( |\tilde{\vt{a}}^T_j \tilde{\vt{z}}| + \epsilon_j \|\vt{z}\|_{q^*} \right) }
\end{align*}
As a result for any $1 \leq j \leq K$ we have
\begin{align*}
|\vt{a}^T_j \vt{\theta}_F | = |(\tilde{\vt{a}}_j + \vt{e}_j )^T \alpha \vt{z} | 
\leq |\alpha| \left( |\tilde{\vt{a}}^T_j \vt{z}| + \epsilon_j \|\vt{z}\|_{q^*} \right) \leq 1.
\end{align*}
This implies that the dual point $\vt{\theta}_F$ is feasible, i.e. $\vt{\theta}_{F} \in \Delta_{\mt{A}}$. 
The fact that $\vt{\theta}_{F} \in \Delta_{\tilde{\mt{A}}}$ follows similarly (since $\epsilon_j \|\vt{z}\|_{q^*} > 0$).

\subsection{Detailed complexities} \label{apx:complexity_screening}

We suppose that the $p^*$-norms of the atoms $\|\vt{a}_j\|_{p^*}$ and 
$q$-norms of the atom approximation error $\|\vt{e}_j\|_q$
are precalculated and stored and memory.
The total screening complexity is summarized in Table \ref{table:complexity_iteration} and detailed in the following.

\begin{table}[h]
\renewcommand{\arraystretch}{1.3}
\centering
	\begin{tabular}{|c|c|c|c|c|}
	\hline
	Screening &$\vt{\theta}_t$ & $\mathcal{R}$ & Test & \textbf{Total} \\ \hline \hline
	Dynamic Safe &$3N +  |\mathcal{A}|$ & $N$ &  $|\mathcal{A}|$ & $4N + 2|\mathcal{A}|$ \\ \hline
	Stable Dynamic Safe &$4N +  2|\mathcal{A}|$ & $N$ &  $|\mathcal{A}|$ & $5N + 3|\mathcal{A}|$ \\ \hline
	GAP Safe &$3N +  |\mathcal{A}|$ & $N$ &  $|\mathcal{A}|$ & $4N + 2|\mathcal{A}|$ \\ \hline
	Stable GAP Safe &$4N +  2|\mathcal{A}|$ & $N$ &  $N+2|\mathcal{A}|$ & $6N + 4|\mathcal{A}|$ \\ \hline
	\end{tabular}
\caption{\label{table:complexity_screening}
Screening: number of floating-point operations.} 
\end{table}

\subsubsection{Feasible point $\pmb{\theta}_t$}
A total of at most $3N +  |\mathcal{A}_t|$ operations for the point in eq. \eqref{eq:theta_dynamic}, distributed as follows:
\begin{description}[leftmargin=!,labelwidth=1cm]
\item[$N$] dot product $\vt{y}^T\vt{\rho}_t$.
\item[$N$] norm $\|\vt{\rho}_t\|^2_2$ (if primal objective not calculated as a convergence criterion).
\item[$N$] product $\alpha \vt{\rho}_t$.
\item[$|\mathcal{A}_t|$] comparisons for the infinity norm $\|\mt{A}^T\vt{\rho}\|_{\infty}$.
\end{description}
At most $N \!+\! |\mathcal{A}_t|$ extra operations for the point in eq. \eqref{eq:theta_dynamic_approx}:
\begin{description}[leftmargin=!,labelwidth=1cm]
\item[$N$] norm $\|\tilde{\vt{\rho}}_t\|_{q^*}$ (only if $q^* \neq 2$).
\item[$|\mathcal{A}_t|$] products $\epsilon_j\|\tilde{\vt{\rho}}_t\|_{q^*}$ on the calculation of $\alpha'_t$.
\end{description}

\

\subsubsection{Safe region $\mathcal{R}$}
At most $N$ operations for computing $\|\vt{\theta}_t - \vt{y}/\lambda\|_2$ (if the dual objective $D(\vt{\theta}_t)$ is not already calculated as a convergence criterion) for the radius of either Dynamic Safe or GAP Safe sphere. 

\

\subsubsection{Screening test}
Other operations (aside from the mentioned $\mt{A}^T\vt{c}$) sum up to $|\mathcal{A}_t|$ operations for the product $R\|\vt{a}_j\|_{p^*} ~ \forall~ j \in \mathcal{A}_t$.
The proposed stable screening tests require at most an extra $N+|\mathcal{A}_t|$ operations:
\begin{description}[leftmargin=!,labelwidth=1cm]
\item[$N$]  norm $\|\vt{c}\|_{q^*}$ (if $q^* \neq 2$ and $\vt{c}$ varies with $t$).
\item[$|\mathcal{A}_t|$]  products $\epsilon_j\|\vt{c}\|_{q^*} ~\forall~ j \in \mathcal{A}_t$  (if $\vt{c}$ varies with $t$).
\end{description}

\bibliographystyle{IEEEtran}
\bibliography{IEEEabrv,./PhD,./Mestrado}

\begin{thebibliography}{10}
\providecommand{\url}[1]{#1}
\csname url@samestyle\endcsname
\providecommand{\newblock}{\relax}
\providecommand{\bibinfo}[2]{#2}
\providecommand{\BIBentrySTDinterwordspacing}{\spaceskip=0pt\relax}
\providecommand{\BIBentryALTinterwordstretchfactor}{4}
\providecommand{\BIBentryALTinterwordspacing}{\spaceskip=\fontdimen2\font plus
\BIBentryALTinterwordstretchfactor\fontdimen3\font minus
  \fontdimen4\font\relax}
\providecommand{\BIBforeignlanguage}[2]{{%
\expandafter\ifx\csname l@#1\endcsname\relax
\typeout{** WARNING: IEEEtran.bst: No hyphenation pattern has been}%
\typeout{** loaded for the language `#1'. Using the pattern for}%
\typeout{** the default language instead.}%
\else
\language=\csname l@#1\endcsname
\fi
#2}}
\providecommand{\BIBdecl}{\relax}
\BIBdecl

\bibitem{F.Dantas2019a}
C.~F.~Dantas and R.~Gribonval, ``Stable screening - {Python} code,''
  \href{https://hal.inria.fr/hal-02129219}{⟨hal-02129219⟩}, May 2019.

\bibitem{Adler2012}
A.~Adler, V.~Emiya, M.~G. Jafari, M.~Elad, R.~Gribonval, and M.~D. Plumbley,
  ``Audio inpainting,'' \emph{{IEEE Transactions on Audio, Speech and Language
  Processing}}, vol.~20, no.~3, pp. 922--932, Mar 2012.

\bibitem{Malioutov2005}
D.~Malioutov, M.~Cetin, and A.~S. Willsky, ``A sparse signal reconstruction
  perspective for source localization with sensor arrays,'' \emph{IEEE
  Transactions on Signal Processing}, vol.~53, no.~8, pp. 3010--3022, Aug 2005.

\bibitem{Dossal2005}
C.~Dossal and S.~Mallat, ``Sparse spike deconvolution with minimum scale,'' in
  \emph{Signal Processing with Adaptive Sparse Structured Representations
  (SPARS workshop)}, Rennes, France, Nov 2005, pp. 123--126.

\bibitem{Mallat1993}
S.~G. Mallat and Z.~Zhang, ``Matching pursuits with time-frequency
  dictionaries,'' \emph{IEEE Transactions on Signal Processing}, vol.~41,
  no.~12, pp. 3397--3415, Dec 1993.

\bibitem{Pati1993}
Y.~C. Pati, R.~Rezaiifar, and P.~S. Krishnaprasad, ``Orthogonal matching
  pursuit: recursive function approximation with applications to wavelet
  decomposition,'' in \emph{27th Asilomar Conference on Signals, Systems and
  Computers}, Nov 1993, pp. 40--44.

\bibitem{Chen1998}
S.~S. Chen, D.~L. Donoho, and M.~A. Saunders, ``Atomic decomposition by basis
  pursuit,'' \emph{SIAM Journal on Scientific Computing}, vol.~20, no.~1, pp.
  33--61, Jan 1998.

\bibitem{Tibshirani1996}
R.~Tibshirani, ``Regression shrinkage and selection via the lasso,''
  \emph{Journal of the Royal Statistical Society. Series B (Methodological)},
  vol.~58, no.~1, pp. 267--288, Jan 1996.

\bibitem{Daubechies2004}
I.~Daubechies, M.~Defrise, and C.~De~Mol, ``An iterative thresholding algorithm
  for linear inverse problems with a sparsity constraint,''
  \emph{Communications on Pure and Applied Mathematics}, vol.~57, no.~11, pp.
  1413--1457, Aug 2004.

\bibitem{Beck2009}
A.~Beck and M.~Teboulle, ``A fast iterative shrinkage-thresholding algorithm
  for linear inverse problems,'' \emph{SIAM Journal on Imaging Sciences},
  vol.~2, no.~1, pp. 183--202, Jan 2009.

\bibitem{Bioucas-Dias2007}
J.~M. Bioucas-Dias and M.~A.~T. Figueiredo, ``A new {TwIST}: Two-step iterative
  shrinkage/thresholding algorithms for image restoration,'' \emph{IEEE
  Transactions on Image Processing}, vol.~16, no.~12, pp. 2992--3004, Dec 2007.

\bibitem{Wright2009}
S.~J. Wright, R.~D. Nowak, and M.~A.~T. Figueiredo, ``Sparse reconstruction by
  separable approximation,'' \emph{IEEE Transactions on Signal Processing},
  vol.~57, no.~7, pp. 2479--2493, Jul 2009.

\bibitem{Chambolle2011}
A.~Chambolle and T.~Pock, ``A first-order primal-dual algorithm for convex
  problems with applications to imaging,'' \emph{Journal of Mathematical
  Imaging and Vision}, vol.~40, no.~1, pp. 120--145, May 2011.

\bibitem{Rubinstein2010a}
R.~Rubinstein, M.~Zibulevsky, and M.~Elad, ``Double sparsity: Learning sparse
  dictionaries for sparse signal approximation,'' \emph{IEEE Transactions on
  Signal Processing}, vol.~58, no.~3, pp. 1553--1564, Mar 2010.

\bibitem{Chabiron2015}
O.~Chabiron, F.~Malgouyres, J.-Y. Tourneret, and N.~Dobigeon, ``Toward fast
  transform learning,'' \emph{International Journal of Computer Vision}, vol.
  114, no.~2, pp. 195--216, Sep 2015.

\bibitem{Magoarou2016}
L.~L. Magoarou and R.~Gribonval, ``Flexible multilayer sparse approximations of
  matrices and applications,'' \emph{IEEE Journal of Selected Topics in Signal
  Processing}, vol.~10, no.~4, pp. 688--700, Jun 2016.

\bibitem{Sulam2016}
J.~Sulam, B.~Ophir, M.~Zibulevsky, and M.~Elad, ``Trainlets: Dictionary
  learning in high dimensions,'' \emph{IEEE Transactions on Signal Processing},
  vol.~64, no.~12, pp. 3180--3193, Jun 2016.

\bibitem{Dantas2017}
C.~F. {Dantas}, M.~N. {da Costa}, and R.~d.~R.~{Lopes}, ``Learning dictionaries
  as a sum of {Kronecker} products,'' \emph{IEEE Signal Processing Letters},
  vol.~24, no.~5, pp. 559--563, May 2017.

\bibitem{ElGhaoui2010}
L.~El~Ghaoui, V.~Viallon, and T.~Rabbani, ``Safe feature elimination in sparse
  supervised learning,'' \emph{EECS Department, University of California,
  Berkeley, Tech. Rep}, Sep 2010.

\bibitem{Xiang2011}
Z.~J. Xiang, H.~Xu, and P.~J. Ramadge, ``Learning sparse representations of
  high dimensional data on large scale dictionaries.'' in \emph{Advances in
  Neural Information Processing Systems}, Dec 2011, pp. 900--908.

\bibitem{Bonnefoy2015}
A.~Bonnefoy, V.~Emiya, L.~Ralaivola, and R.~Gribonval, ``Dynamic screening:
  Accelerating first-order algorithms for the lasso and group-lasso,''
  \emph{IEEE Transactions on Signal Processing}, vol.~63, no.~19, pp.
  5121--5132, Oct 2015.

\bibitem{Fercoq2015}
O.~Fercoq, A.~Gramfort, and J.~Salmon, ``Mind the duality gap: safer rules for
  the lasso,'' in \emph{International Conference on Machine Learning}, vol.~37,
  Jul 2015, pp. 333--342.

\bibitem{Ndiaye2017}
E.~Ndiaye, O.~Fercoq, A.~Gramfort, and J.~Salmon, ``Gap safe screening rules
  for sparsity enforcing penalties,'' \emph{Journal of Machine Learning
  Research}, vol.~18, no. 128, pp. 1--33, Nov 2017.

\bibitem{Xiang2016}
Z.~J. {Xiang}, Y.~{Wang}, and P.~J. {Ramadge}, ``Screening tests for lasso
  problems,'' \emph{IEEE Transactions on Pattern Analysis and Machine
  Intelligence}, vol.~39, no.~5, pp. 1008--1027, May 2017.

\bibitem{F.Dantas2017}
C.~F.~Dantas and R.~Gribonval, ``Dynamic screening with approximate
  dictionaries,'' in \emph{{Colloque GRETSI}}, Juan-les-Pins, France, Sep 2017.

\bibitem{F.Dantas2018}
------, ``Faster and still safe: combining screening techniques and structured
  dictionaries to accelerate the lasso,'' in \emph{IEEE International
  Conference on Acoustics, Speech and Signal Processing}, Calgary, AB, Canada,
  Apr 2018, pp. 4069--4073.

\bibitem{Foucart2013}
S.~Foucart and H.~Rauhut, \emph{A Mathematical Introduction to Compressive
  Sensing}, 2013.

\bibitem{Hawe2013}
S.~Hawe, M.~Seibert, and M.~Kleinsteuber, ``Separable dictionary learning,'' in
  \emph{IEEE Conference on Computer Vision and Pattern Recognition}, Jun 2013,
  pp. 438--445.

\bibitem{F.Dantas2018a}
C.~F.~Dantas, J.~E. Cohen, and R.~Gribonval, ``Learning fast dictionaries for
  sparse representations using low-rank tensor decompositions,'' in
  \emph{International Conference on Latent Variable Analysis and Signal
  Separation}, Guildford, United Kingdom, Jul 2018, pp. 456--466.

\bibitem{Fan2008}
J.~Fan and J.~Lv, ``Sure independence screening for ultrahigh dimensional
  feature space,'' \emph{Journal of the Royal Statistical Society: Series B
  (Statistical Methodology)}, vol.~70, no.~5, pp. 849--911, Nov 2008.

\bibitem{Tibshirani2011}
R.~Tibshirani, J.~Bien, J.~Friedman, T.~Hastie, N.~Simon, J.~Taylor, and R.~J.
  Tibshirani, ``Strong rules for discarding predictors in lasso-type
  problems,'' \emph{Journal of the Royal Statistical Society: Series B
  (Statistical Methodology)}, vol.~74, no.~2, pp. 245--266, Nov 2011.

\bibitem{Xiang2012}
Z.~J. Xiang and P.~J. Ramadge, ``Fast lasso screening tests based on
  correlations,'' in \emph{IEEE International Conference on Acoustics, Speech
  and Signal Processing}, Mar 2012, pp. 2137--2140.

\bibitem{Wang-Wonka2015}
J.~Wang, P.~Wonka, and J.~Ye, ``Lasso screening rules via dual polytope
  projection,'' \emph{Journal of Machine Learning Research}, vol.~16, no.~1,
  pp. 1063--1101, May 2015.

\bibitem{Malti2016}
A.~Malti and C.~Herzet, ``Safe screening tests for lasso based on firmly
  non-expansiveness,'' in \emph{IEEE International Conference on Acoustics,
  Speech and Signal Processing}, Mar 2016, pp. 4732--4736.

\bibitem{Liu2014}
J.~Liu, Z.~Zhao, J.~Wang, and J.~Ye, ``Safe screening with variational
  inequalities and its application to lasso,'' in \emph{International
  Conference on Machine Learning}, vol. 32(2), Jun 2014, pp. 289--297.

\bibitem{Kim2010}
J.~Kim and H.~Park, ``Fast active-set-type algorithms for l1-regularized linear
  regression,'' in \emph{Proceedings of the Thirteenth International Conference
  on Artificial Intelligence and Statistics}, May 2010, pp. 397--404.

\bibitem{Johnson2015}
T.~Johnson and C.~Guestrin, ``Blitz: A principled meta-algorithm for scaling
  sparse optimization,'' in \emph{International Conference on Machine
  Learning}, Lille, France, Jul 2015, pp. 1171--1179.

\bibitem{Massias2017}
A.~G. M.~Massias and J.~Salmon, ``From safe screening rules to working sets for
  faster lasso-type solvers,'' in \emph{NIPS Workshop on Optimization for
  Machine Learning}, Long Beach, USA, Dec 2017.

\bibitem{Massias2018}
M.~Massias, J.~Salmon, and A.~Gramfort, ``Celer: a fast solver for the lasso
  with dual extrapolation,'' in \emph{International Conference on Machine
  Learning}, Stockholm, Sweden, Jul 2018, pp. 3321--3330.

\bibitem{Herzet2018}
C.~Herzet and A.~Dr{\'e}meau, ``Joint screening tests for lasso,'' in
  \emph{IEEE International Conference on Acoustic, Speech and Signal
  Processing}, Calgary, AB, Canada, Apr 2018, pp. 4084--4088.

\bibitem{Osborne2000}
M.~R. Osborne, B.~Presnell, and B.~A. Turlach, ``On the lasso and its dual,''
  \emph{Journal of Computational and Graphical Statistics}, vol.~9, no.~2, pp.
  319--337, Jun 2000.

\bibitem{Matsuura1995}
K.~Matsuura and Y.~Okabe, ``Selective minimum-norm solution of the biomagnetic
  inverse problem,'' \emph{IEEE Transactions on Biomedical Engineering},
  vol.~42, no.~6, pp. 608--615, Jun 1995.

\bibitem{Gramfort2012}
A.~Gramfort, M.~Kowalski, and M.~H{\"a}m{\"a}l{\"a}inen, ``{Mixed-norm
  estimates for the M/EEG inverse problem using accelerated gradient
  methods},'' \emph{Physics in Medicine \& Biology}, vol.~57, no.~7, pp.
  1937--1961, Mar 2012.

\bibitem{Gramfort2014}
A.~Gramfort, M.~Luessi, E.~Larson, D.~A. Engemann, D.~Strohmeier, C.~Brodbeck,
  L.~Parkkonen, and M.~S. H{\"a}m{\"a}l{\"a}inen, ``{MNE software for
  processing MEG and EEG data},'' \emph{Neuroimage}, vol.~86, pp. 446--460, Feb
  2014.

\bibitem{Bredies2013}
K.~Bredies and H.~K. Pikkarainen, ``Inverse problems in spaces of measures,''
  \emph{ESAIM: Control, Optimisation and Calculus of Variations}, vol.~19,
  no.~1, pp. 190--218, Mar 2013.

\end{thebibliography}

\end{document}